\documentclass[11pt]{article}

\usepackage[final]{acl}

\usepackage{times}
\usepackage{latexsym}

\usepackage[T1]{fontenc}

\usepackage[utf8]{inputenc}

\usepackage{microtype}
\usepackage{inconsolata}

\usepackage{graphicx}
\usepackage{hyperref}
\usepackage{fontawesome5}
\usepackage{natbib}
\usepackage{amsmath}
\usepackage{booktabs}  
\usepackage{pifont}    
\usepackage{array}     
\usepackage{subcaption} 
\usepackage{diagbox}
\usepackage{multirow}
\usepackage{float}
\usepackage{multicol} 
\usepackage{geometry}
\usepackage[most]{tcolorbox}
\usepackage{listings}
\usepackage{graphicx}
\usepackage{amssymb} 
\usepackage{tcolorbox}
\usepackage[table]{xcolor}
\usepackage{colortbl} 
\usepackage{tabularx}
\usepackage{longtable}
\usepackage{url}
\usepackage{enumitem}
\tcbuselibrary{skins, breakable} 

\newtcolorbox{finalreportbox}[2][]{
    colback=white,                
    colframe=teal!45!black,       
    coltitle=white,
    colbacktitle=teal!45!black,
    fonttitle=\bfseries\Large,    
    title={#2},
    arc=2mm,
    boxrule=1pt,
    enhanced jigsaw,              
    drop shadow,
    breakable,                    
    lines before break=2,         
    pad at break=2mm,             
    #1
}

\newtcolorbox{evalbox}[2][]{
    colback=white,                
    colframe=gray!40!black,       
    coltitle=white,               
    colbacktitle=gray!40!black,   
    fonttitle=\bfseries\large,    
    title={#2},                   
    arc=2mm,                      
    boxrule=1pt,                  
    enhanced,                     
    drop shadow,                  
    #1
}

\newtcolorbox{notebookbox}[2]{
	enhanced,
	breakable,
	colback=white,
	colframe=#2!75!black,
	coltitle=white,
	fonttitle=\bfseries\large,
	title={#1},
	boxrule=0.8pt,
	arc=2mm,
	toptitle=2mm,
	bottomtitle=2mm,
	left=2mm, right=2mm, top=3mm, bottom=3mm,
	before skip=10pt, after skip=10pt
}

\usepackage{listings}
\usepackage{xcolor} 

\definecolor{codegreen}{rgb}{0,0.6,0}
\definecolor{codegray}{rgb}{0.5,0.5,0.5}
\definecolor{codepurple}{rgb}{0.58,0,0.82}
\definecolor{backcolour}{rgb}{0.95,0.95,0.92}

\lstdefinestyle{jupytercode}{
    backgroundcolor=\color{backcolour},   
    commentstyle=\color{codegreen},       
    keywordstyle=\color{magenta},         
    numberstyle=\tiny\color{codegray},    
    stringstyle=\color{codepurple},       
    basicstyle=\ttfamily\footnotesize,    
    breakatwhitespace=false,         
    breaklines=true,                      
    captionpos=b,                    
    keepspaces=true,                 
    numbers=left,                         
    numbersep=5pt,                  
    showspaces=false,                
    showstringspaces=false,
    showtabs=false,                  
    tabsize=2,
    frame=single,                         
    rulecolor=\color{lightgray!50},       
    postbreak=\mbox{\textcolor{red}{$\hookrightarrow$}\space},
    xleftmargin=0pt,        
    framexleftmargin=0pt,   
    framesep=3pt,           
    resetmargins=true       
    ,gobble=8               
}

\lstdefinestyle{jupyteroutput}{
    basicstyle=\ttfamily\scriptsize,      
    breaklines=true,                      
    breakatwhitespace=false,
    frame=none,                           
    backgroundcolor=\color{white},        
    columns=fullflexible,                 
    keepspaces=true,
    gobble=8
}
\newcommand{\inlabel}[1]{%
	\par\noindent\hspace*{0.25em}{\footnotesize\color{blue!60!black}\ttfamily In [#1]:}%
	\par\nobreak\vspace{-0.35\baselineskip}%
}
\newcommand{\outlabel}[1]{%
	\par\noindent\hspace*{0.25em}{\footnotesize\color{red!60!black}\ttfamily Out [#1]:}%
	\par\nobreak\vspace{-0.35\baselineskip}%
}
\usepackage[most]{tcolorbox}
\usepackage[T1]{fontenc} 

\newtcolorbox{promptbox}[2]{
	enhanced,
	colback=#2!5!white,
	colframe=#2!80!black,
	coltitle=white,
	fonttitle=\bfseries,
	title={#1},
	boxrule=0.8pt,
	arc=2mm,
	outer arc=2mm,
	left=3mm, right=3mm,
	top=3mm, bottom=3mm,
	toptitle=2mm,
	bottomtitle=2mm,
	before skip=10pt,
	after skip=10pt,
	breakable 
}

\newcommand{\cmark}{\textcolor{green!70!black}{\ding{51}}}
\newcommand{\xmark}{\textcolor{red!80!black}{\ding{55}}}

\title{DSAEval: Evaluating Data Science Agents on a Wide Range of Real-World Data Science Problems}

\author{
  \textbf{Maojun Sun}\textsuperscript{1,\textdagger} \quad
  \textbf{Yifei Xie}\textsuperscript{1,\textdagger} \quad
  \textbf{Yue Wu}\textsuperscript{1,\textdagger} \quad
  \textbf{Ruijian Han}\textsuperscript{1,*} \\
  \textbf{Binyan Jiang}\textsuperscript{1} \quad
  \textbf{Defeng Sun}\textsuperscript{2} \quad
  \textbf{Yancheng Yuan}\textsuperscript{2,*} \quad
  \textbf{Jian Huang}\textsuperscript{1,2,*} \\
  \\
  \textsuperscript{1}Department of Data Science and Artificial Intelligence,
  The Hong Kong Polytechnic University \\
  \textsuperscript{2}Department of Applied Mathematics,
  The Hong Kong Polytechnic University
}

\begin{document}
\maketitle
\begingroup
\renewcommand{\thefootnote}{\fnsymbol{footnote}}
\footnotetext[1]{Corresponding authors:\\
\mbox{\scriptsize\texttt{\{ruijian.han, yancheng.yuan, j.huang\}@polyu.edu.hk}}.}
\footnotetext[2]{These authors contributed equally to this work.}
\endgroup


\begin{abstract}
Recent LLM-based data agents aim to automate data science tasks ranging from data analysis to deep learning. 
However, the open-ended nature of real-world data science problems, which often span multiple taxonomies and lack standard answers, poses a significant challenge for evaluation. 
To address this, we introduce DSAEval, a benchmark comprising 641 real-world data science problems grounded in 285 diverse datasets, covering both structured and unstructured data (e.g., image and text).
DSAEval incorporates three distinctive features: 
(1) Multimodal Environment Perception, which enables agents to interpret observations from multiple modalities, including text and vision; 
(2) Multi-Query Interactions, which mirror the iterative and cumulative nature of real-world data science projects; and 
(3) Multi-Dimensional Evaluation, which provides a holistic assessment across reasoning, code, and results. 
We systematically evaluate 13 recent advanced agentic LLMs using DSAEval.
Our results show that Claude-Sonnet-4.5 achieves the strongest overall performance, MiMo-V2-Pro and GPT-5.2 lead in duration and step efficiency, respectively, and MiMo-V2-Flash is the most cost-effective. We further demonstrate that multimodal perception consistently improves performance on vision-related tasks, with gains ranging from 2.04\% to 11.30\%. Overall, while current data science agents perform well on structured data and routine data analysis workflows, substantial challenges remain in unstructured domains. Finally, we offer critical insights and outline future research directions. 



\begin{center}
\faGithub\ \href{https://ama-cmfai.github.io/DSAEval/}{https://ama-cmfai.github.io/DSAEval/}
\end{center}
\end{abstract}
\section{Introduction}

\label{subsec:main_results}
\begin{figure*}[t]
    \centering
    \includegraphics[width=\textwidth]{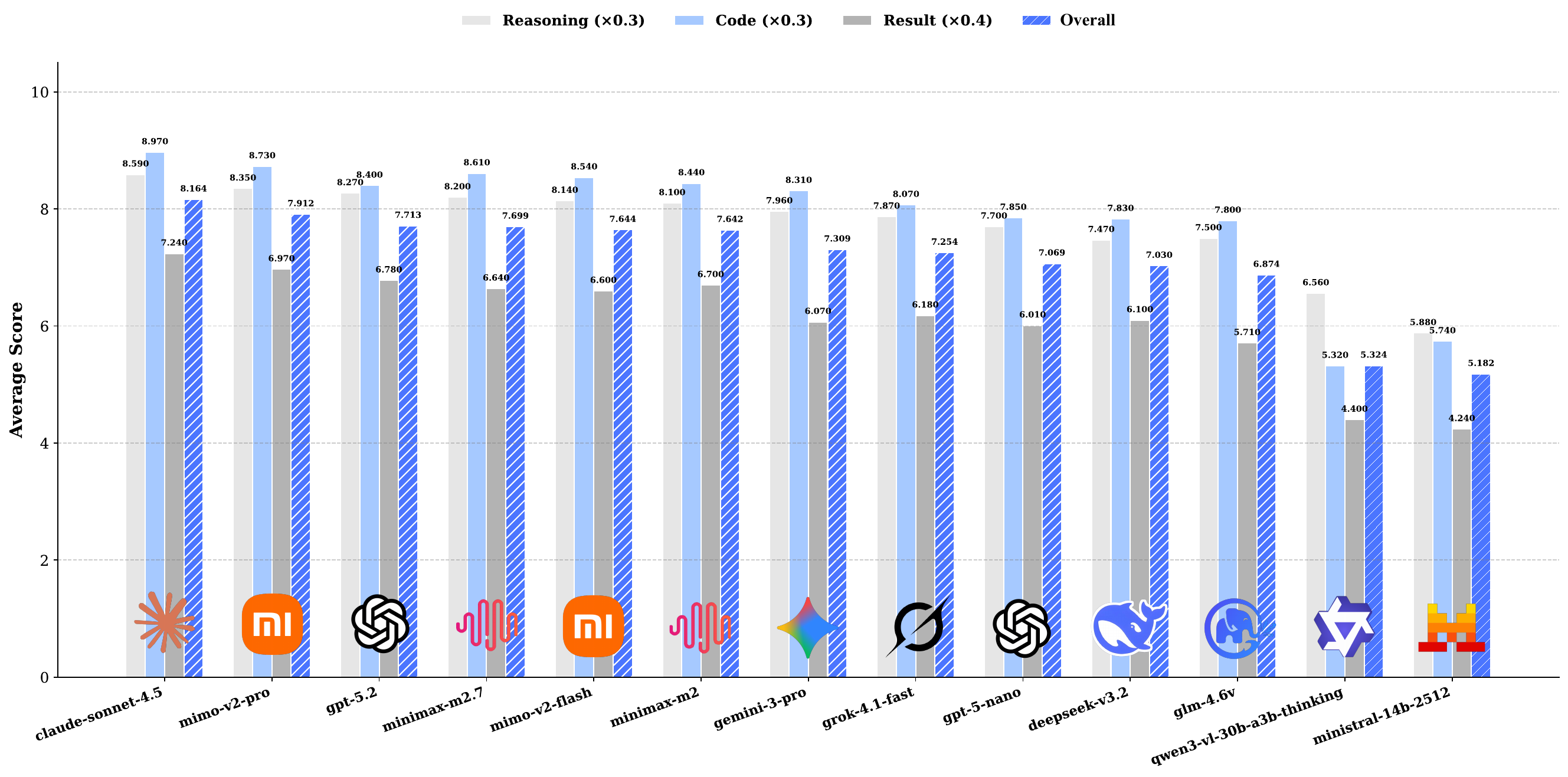}
    \caption{Overall performance of all selected models on DSAEval.}
    \label{fig:overall_perf}
\end{figure*}

Recent advances in Large Language Model (LLM)-based data science agents (or data agents) have significantly promoted the automation of data science, encompassing a wide range of tasks from exploratory data analysis and traditional machine learning to complex deep learning workflows \citep{sun2025survey, hong2024datainterpreterllmagent, zhang2025deepanalyze, sun2026rejoinder2, zhang2023data, sun2026dare}. Nevertheless, evaluating the efficacy of these agents remains a formidable challenge \citep{dscodebench, sun2025survey, sun2026rejoinder}. Real-world data science problems are inherently open-ended and exploratory  \citep{nascimento2024llm4dsevaluatinglargelanguage, hu2024infiagentdabenchevaluatingagentsdata}, often lacking unique, standardized solutions, which renders traditional exact-match evaluation metrics insufficient. Moreover, data science is a multifaceted discipline with numerous specialized domains and intricate workflow stages, necessitating a granular and comprehensive evaluation \citep{datascibench,dsbench}. However, most prior benchmarks have been limited in scope, focusing only on partial sub-domains or isolated stages in the pipeline. Moreover, traditional evaluation protocols often rely on objective ground truth that is easily verifiable, such as code unit tests \citep{ds1000, dscodebench}, metric comparisons \citep{huang2024dacodeagentdatascience, huang2024mlagentbenchevaluatinglanguageagents, dsbench}, or exact matches \citep{hu2024infiagentdabenchevaluatingagentsdata, egg2025dabstepdataagentbenchmark}. In most real-world data analysis scenarios, however, the final deliverable is not a single deterministic answer but a report-style output that synthesizes quantitative results, visual evidence, methodological choices, and interpretive conclusions. This open-ended nature makes it difficult for conventional evaluation methods to fully assess the quality of data science agents. To address these dilemmas, we introduce DSAEval, a comprehensive benchmark designed to evaluate data science agents using large-scale, real-world problems. DSAEval spans broad domains of data science problems, including Statistical Testing \& Inference (STI), Data Analysis (DA), Time Series (TS) Analysis, Natural Language Processing (NLP), Computer Vision (CV), and Clustering. Furthermore, it incorporates critical workflow stages such as Data Ingestion, Feature Engineering, Modeling, and Inference. 

To simulate authentic usage scenarios, DSAEval incorporates four distinctive features:
(1) \textbf{Broad Domain Coverage}: DSAEval covers fine-grained data science domains, with substantial deep learning tasks supported by a GPU-equipped sandbox.
(2) \textbf{Multimodal Environment Perception}: We empower agents to interpret and interact with diverse modalities within the environment, such as text, tabular data, and visual plots.
(3) \textbf{Multi-Query Interactions}: Each session is grounded in a dataset and consists of consecutive, interdependent tasks, reflecting the iterative nature of real data science workflows.
(4) \textbf{Multi-Dimensional Evaluation}: To provide a holistic assessment of the entire pipeline, we employ tailored LLM judges to evaluate agent performance across multiple dimensions, including reasoning quality, code correctness, and the correctness and clarity of the final report. We further validate the reliability of these judges through a Human-LLM Alignment Study and an inter-judge agreement study. DSAEval evaluates end-to-end data science problem solving over provided datasets: agents address real-world analytical queries in a persistent, GPU-enabled sandbox and produce executable notebooks and analytical reports containing key numerical results or conclusions.

We conducted a systematic evaluation of 13 recent LLMs and Vision-Language Models (VLMs) across multiple dimensions, including overall performance, efficiency, and cost-effectiveness. Overall, Claude-Sonnet-4.5 achieves the best performance on DSAEval (Figure~\ref{fig:overall_perf}), while MiMo-V2-Pro, GPT-5.2, and MiMo-V2-Flash demonstrate strong duration efficiency, token/step efficiency, and cost-effectiveness, respectively. Notably, compared with the single-text-observation baseline, the proposed Multimodal Environment Perception mechanism leads to consistent performance improvements across diverse vision-related tasks, with gains ranging from 2.04\% to 11.30\%. Our empirical results further show that current models are highly proficient in processing structured tabular data and executing standard data engineering tasks, but they still face substantial challenges in unstructured domains, such as Computer Vision and NLP, as well as in unsupervised learning tasks such as Clustering. Based on these findings, we provide critical insights and outline future research directions for advancing data science agents and their evaluation. 


\section{Background and Related Works}

\begin{table*}[h]
\centering
\caption{Comparison of Agentic Benchmarks for Data Science. 
\textbf{Datasets}: The number of real-world datasets, where `-' indicates that the exact number is unknown and \xmark indicates that no datasets are included.
\textbf{Hetero. Data}: Contains heterogeneous data sources (e.g., Tabular, Images).
\textbf{Vision-modal Obs}: Supports vision modality observations. 
\textbf{Multi-step}: Requires multi-step reasoning.
\textbf{Deep Learning}: Includes deep learning tasks.
\textbf{Eval. Method}: The evaluation method of the study. CUT = Code Unit Tests, EM = Exact/Structured Matching Against Ground Truth, MC = Metric Comparison, MCQ = Multiple Choice Question.}
\label{tab:comparison}
\resizebox{\textwidth}{!}{%
\begin{tabular}{lccccccc}
\toprule
\textbf{Benchmark} & \textbf{Datasets} & \textbf{Questions} & \textbf{Hetero. Data} & \textbf{Vision-modal Obs} & \textbf{Multi-step} & \textbf{Deep Learning} & \textbf{Eval. Format} \\
\midrule
DS-1000 \citep{ds1000} & \xmark & 1,000 & \xmark & \xmark & \xmark & \xmark & Closed (CUT) \\
Infiagent-DABench \citep{hu2024infiagentdabenchevaluatingagentsdata} & 52 & 257 & \xmark & \xmark & \cmark & \xmark & Closed (EM) \\
DA-Code \citep{huang2024dacodeagentdatascience} & 500 & 500 & \cmark & \xmark & \cmark & \xmark & Closed (MC) \\
MLAgentBench \cite{huang2024mlagentbenchevaluatinglanguageagents} & 13 & 13 & \cmark & \xmark & \cmark & \cmark & Closed (MC) \\
MLE-Bench \citep{chan2025mle} & 75 & 75 & \cmark & \xmark & \cmark & \cmark & Closed (MC) \\
DSEval \citep{zhang2024benchmarkingdatascienceagents} & 294 & 825 & \xmark & \xmark & \xmark & \xmark & Closed (EM) \\
DSCodeBench \citep{dscodebench} & \xmark & 1,000 & \xmark & \xmark & \xmark & \xmark & Closed (CUT) \\ 
DSBench \citep{dsbench} & 112 & 540 & \cmark & \xmark & \cmark & \cmark & Closed (MCQ, MC) \\
DABstep \citep{egg2025dabstepdataagentbenchmark} & - & 450 & \cmark & \xmark & \cmark & \xmark & Closed (EM) \\

\midrule
\textbf{DSAEval (Ours)} & 285 & 641 & \cmark & \cmark & \cmark & \cmark  & Open (Reason, Code, Report) \\
\bottomrule
\end{tabular}%
}
\end{table*}

\noindent \textbf{LLM-Based Data Science Agents and Benchmarks.}
Recent LLM-based data science agents, such as DataInterpreter \citep{hong2024datainterpreterllmagent}, LAMBDA \citep{sun2025lambda}, DeepAnalyze \citep{zhang2025deepanalyze}, and Data Copilot \citep{zhang2023data}, have shown strong potential in automating data analysis, machine learning, and end-to-end data science workflows.
To evaluate these systems, recent benchmarks such as DS-1000 \citep{ds1000}, DA-Code \citep{huang2024dacodeagentdatascience}, InfiAgent-DABench \citep{hu2024infiagentdabenchevaluatingagentsdata}, MLAgentBench \citep{huang2024mlagentbenchevaluatinglanguageagents}, DSEval \citep{zhang2024benchmarkingdatascienceagents}, DABstep \citep{egg2025dabstepdataagentbenchmark}, DSCodeBench \citep{dscodebench}, and DSBench \citep{dsbench} have provided important testbeds for assessing data science capabilities.
However, most existing benchmarks still focus primarily on tabular data or isolated workflow stages, with limited coverage of complex deep learning tasks.
They also provide limited support for multimodal environment perception, making it difficult to evaluate whether agents can interpret plots, visualizations, image data, and other visual feedback generated during data science workflows.

Moreover, traditional evaluation protocols often rely on objective ground truth that is easy to verify, such as code unit tests \citep{ds1000,dscodebench}, metric comparisons \citep{huang2024dacodeagentdatascience,huang2024mlagentbenchevaluatinglanguageagents,dsbench}, or exact matches \citep{hu2024infiagentdabenchevaluatingagentsdata,egg2025dabstepdataagentbenchmark}.
In real-world data analysis, however, the final deliverable is often a report-style output that integrates code execution, quantitative results, visual evidence, methodological choices, and interpretive conclusions.
Such open-ended outputs are difficult to assess with conventional evaluation methods.
In contrast, DSAEval integrates heterogeneous real-world datasets across diverse domains, includes complex deep learning workflows, supports multimodal observations, and introduces LLM-based evaluators to assess agents across reasoning, code, and final results.
Table~\ref{tab:comparison} provides a detailed comparison with selected related works.

\begin{figure*}[h]
  \centering
  \includegraphics[width=\textwidth]{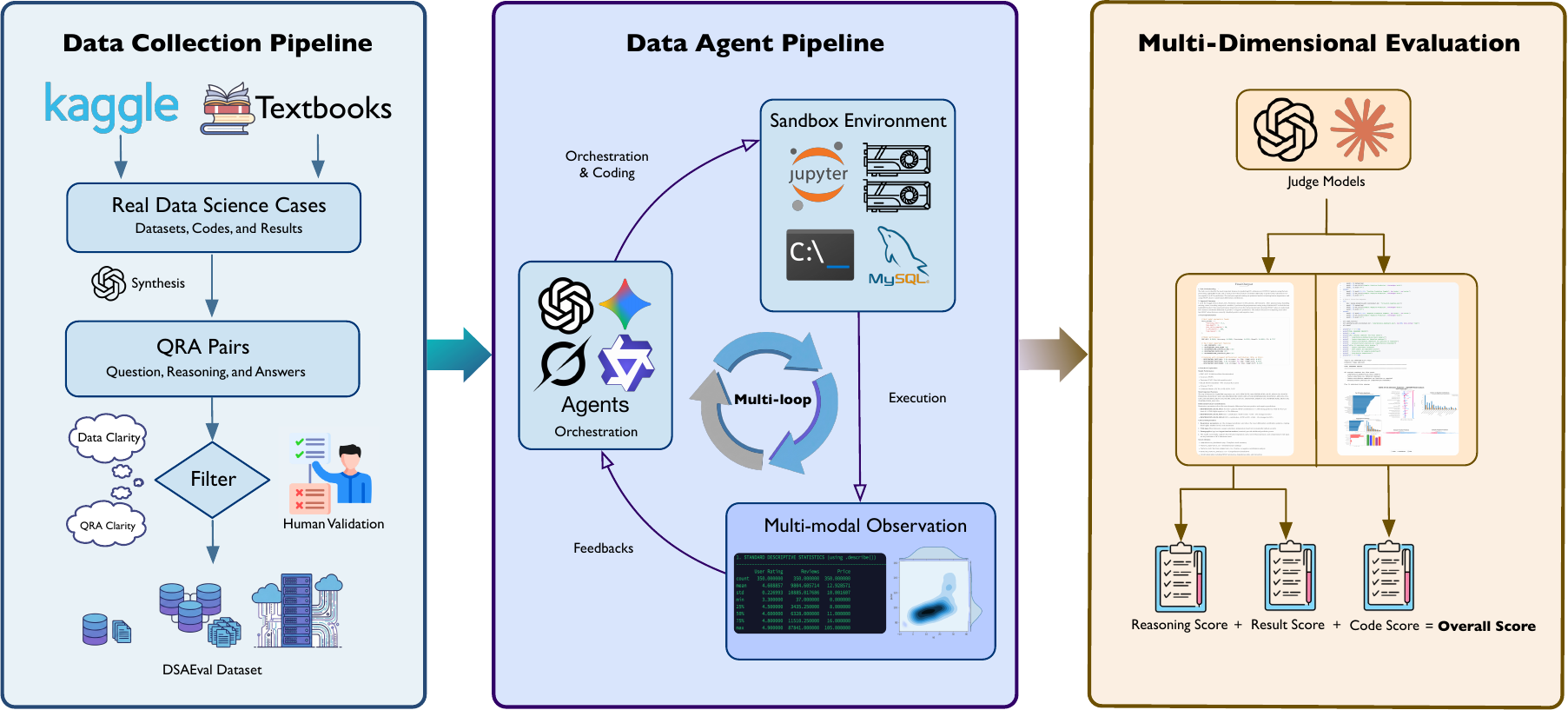}
  \caption{Overview of DSAEval. \textbf{Left:} In the Data Collection Pipeline, raw cases are cleaned and synthesized into Question, Reasoning, and Answer (QRA) pairs using advanced LLMs. \textbf{Middle:} The Data Agent Pipeline orchestrates the agent to solve tasks within a Sandbox Environment. The agent receives multimodal observations and produces a final report and a Jupyter notebook. \textbf{Right:} The Multi-Dimensional Evaluation module employs multiple judge models to score the reasoning, code, and results against the soft ground truth, yielding an overall score.}\label{fig:method}
\end{figure*}

\begin{figure*}[h]
  \centering
  \includegraphics[width=\textwidth]{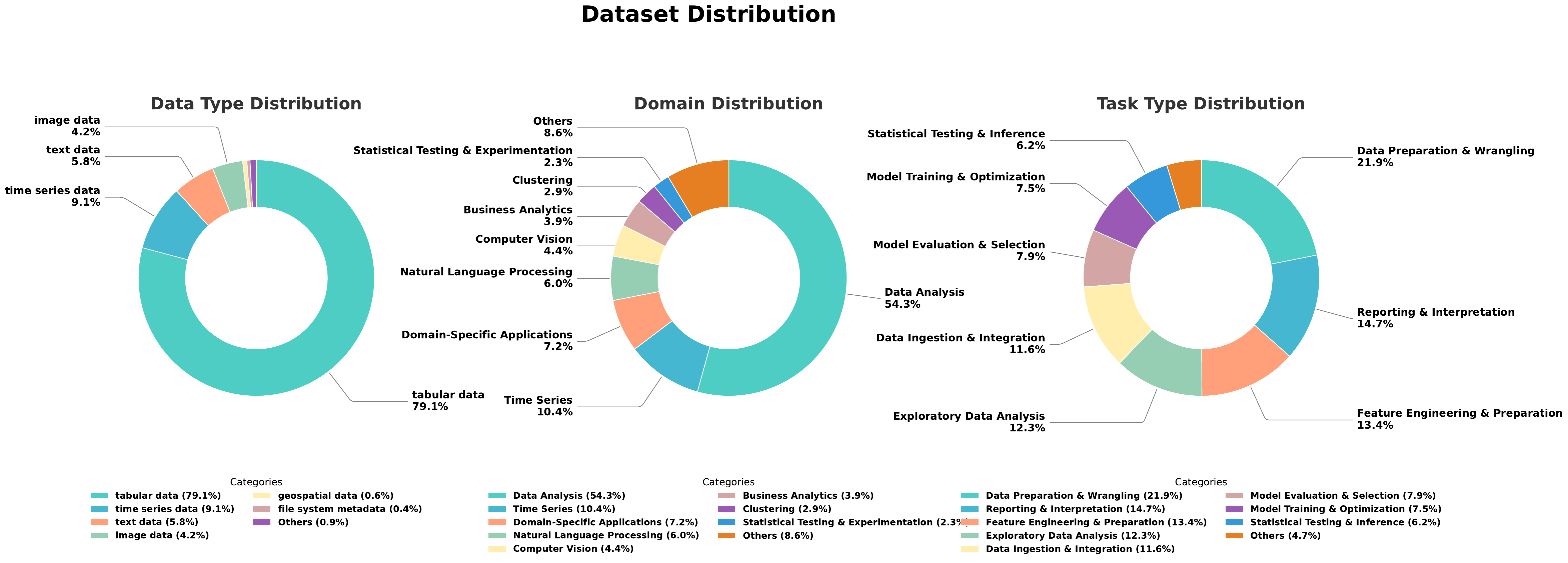}
  \caption{Distribution of the DSAEval benchmark. The suite covers diverse data modalities (left), problem domains (center), and task types (right), ensuring comprehensive evaluation coverage.\label{fig:dataset_dist}}
\end{figure*}

\section{Methodology}
\label{sec:methodology}

To rigorously evaluate the capabilities of data science agents in real-world scenarios, we propose DSAEval, a comprehensive benchmarking framework. As illustrated in Figure \ref{fig:method}, our framework consists of three core components: (1) a rigorously curated benchmark suite derived from real-world data science cases; (2) a sandbox-based interactive environment supporting multimodal perception and continuous multi-query workflows; and (3) a multi-dimensional evaluation protocol utilizing an LLM-based judge to assess the quality of reasoning, code, and result (or report).

\subsection{Benchmark Suite Construction}

To ensure diversity and realism, we constructed a large-scale data pipeline from over 2,000 open-source data science datasets and competitions (e.g., Kaggle) and 50 authoritative data science and statistics textbooks; details are provided in Appendix~\ref{append_data}. These sources cover domain-specific real-world cases ranging from Statistical Inference to Computer Vision. For Kaggle datasets, we selected the highest-voted notebooks to ensure code quality.

We then applied a multi-stage filtering and annotation process. Low-quality cases were removed, including short solutions with fewer than 10 cells, notebooks without clear answers, duplicate topics, and cases depending on unavailable external data. Next, we used advanced LLMs, including GPT-5 \citep{openai_gpt52_2025} and Grok-4 \citep{grok_4}, to synthesize questions, reasoning processes, and corresponding answers, which serve as the reference \textit{soft ground truth} ($G$); the prompt is shown in Appendix~\ref{sec:detail_prompt}. We further curated a representative subset based on data, domain, and task diversity, followed by expert human validation. In total, 1,780 candidates (73.5\%) were rejected and 641 (26.5\%) retained. Appendix~\ref{app:data_validation} provides the rubric and accepted/rejected examples. As shown in Figure~\ref{fig:dataset_dist}, the final benchmark contains 285 heterogeneous datasets and 641 problems, covering domains such as CV, NLP, and TS, and tasks such as Statistical Testing \& Inference, Feature Engineering \& Preparation, and Model Training \& Optimization.

\subsection{Data Agent Framework}
DSAEval introduces a framework designed to mimic authentic human data science workflows.

\paragraph{Multimodal Environment Perception.}
Real-world data science is inherently multimodal. To support this, we incorporated multimodal observations, with a particular emphasis on data visualization. We designed a Sandbox Environment ($\mathcal{E}$) equipped with a Jupyter Notebook kernel and dedicated GPU acceleration. Let $c_t$ denote the executable Python code generated by the agent at step $t$. The environment executes this code and returns a multimodal observation $o_t = \mathcal{E}(c_t)$. The observation $o_t$ is parsed into a structured format containing three modalities of $o_t = \{ o_t^{\text{txt}}, o_t^{\text{tab}}, o_t^{\text{img}} \}$
where $o_t^{\text{txt}}$ represents standard output/error logs, $o_t^{\text{tab}}$ denotes serialized markdown tables or dataframe previews, and $o_t^{\text{img}}$ captures generated plots (e.g., matplotlib figures) as image encodings. This design ensures that the agent can perceive and reason over visual feedback, such as identifying non-linear patterns in a scatter plot or diagnosing training loss curves, thereby maximizing the capability of modern agentic VLMs.

\paragraph{Multi-Query Interactions.}
Complex data science problems require a sequence of interdependent actions. DSAEval models this process as a session $S$. Formally, a session is defined as $S = \{D, Q, H_0\}$, where $D$ is the dataset, $Q = \{q_1, q_2, ..., q_n\}$ is a sequence of logically connected sub-tasks, and $H_0$ is the initial context. For a specific query $q_k \in Q$, the agent $\mathcal{A}$ operates in a loop to generate a thought process and code $c_t$ based on the current history $H_t$ and previous observation $o_{t-1}$:
\begin{equation*}
    (c_t, \text{thought}_t) \leftarrow \mathcal{A}(q_k, H_t, o_{t-1}).
\end{equation*}
The agent system maintains the persistent state of the sandbox kernel throughout the session (e.g., variables created by previously executed code). Upon completing the tasks, the system produces two final artifacts: a complete Code Notebook ($\mathcal{N}$) containing all executed cells, and a final Textual Report ($\mathcal{R}$) summarizing core reasoning steps, code, and the answers of the query.

\subsection{Multi-Dimensional Evaluation Protocol}
Evaluating open-ended data science tasks requires metrics beyond simple exact-match accuracy. Motivated by LLM-as-Judges, we use a Multi-Dimensional Evaluation protocol driven by a specialized "Judge Model" $\mathcal{J}$. The Judge Model assesses the agent's final artifacts ($\mathcal{N}, \mathcal{R}$) against the soft ground truth $G$. The overall score $S_{\text{overall}}$ is calculated as a weighted sum of three distinct dimensions:
\begin{equation*}
    S_{\text{overall}} = \alpha \cdot S_{\text{reason}} + \beta \cdot S_{\text{code}} + (1 - \alpha - \beta) \cdot S_{\text{result}}. 
\end{equation*}

In our experiments, we configure $\alpha = 0.3$ and $\beta = 0.3$. This configuration places a slightly higher emphasis on the accuracy of the final findings ($S_{\text{result}}$) while maintaining substantial importance on the process validity. The components are defined as follows:

\begin{itemize}[noitemsep]
    \item \textbf{Reasoning Process ($S_{\text{reason}}$):} This metric evaluates the conceptual soundness and methodological validity of the agent's approach. The Judge assesses whether the selected statistical or machine learning techniques are appropriate for the problem type (e.g., correct model selection, valid assumptions) and if the logical flow adheres to core data science principles, independent of the final numerical accuracy.
    \item \textbf{Code Steps ($S_{\text{code}}$):} This metric assesses the completeness and correctness of the generated code implementation from the notebook. The evaluation focuses on whether the code sequence is logically coherent, free from fatal errors in key steps, and successfully produces the necessary intermediate outputs required to support the final conclusion.
    \item \textbf{Final Result ($S_{\text{result}}$):} This metric evaluates the holistic quality of the outcome, encompassing quantitative metrics, qualitative insights, and visualizations. Crucially, this evaluation is open-ended: it accepts alternative solutions that differ from the reference but offer valid or superior methodology. The detailed judge-model prompt can be found in Appendix \ref{app:judeg_pmt}.
\end{itemize}

We additionally conduct a sensitivity analysis to examine the robustness of model rankings under different weighting configurations.

\begin{figure*}[h] 
  \centering
  \includegraphics[width=0.95\textwidth]{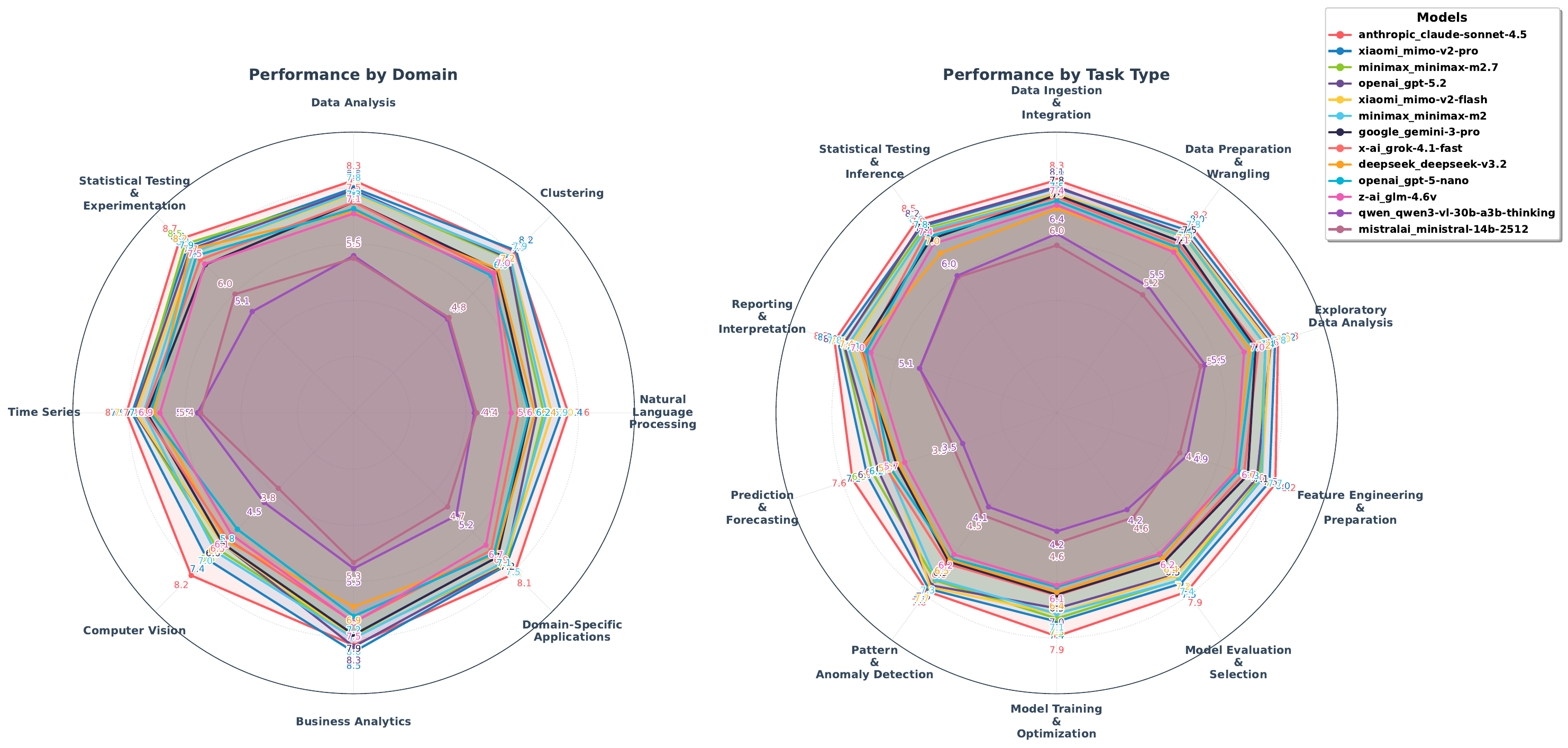}
  \caption{Fine-grained performance analysis by domain and task type.}
  \label{fig:radar_analysis}
\end{figure*}

%
%

\begin{table*}[h]
	\centering
	\caption{Agent performance under single-text and multimodal observations. Parentheses show relative gains; DA: Data Analysis; CV: Computer Vision; PAD: Pattern \& Anomaly Detection; EDA: Exploratory Data Analysis. CIs denote 90\% paired-bootstrap confidence intervals for the absolute score difference, rounded to one decimal.}
	\label{tab:obs_comparison}
	\resizebox{\textwidth}{!}{
		\begin{tabular}{lcccc|cc|cc|cc|cc} 
			\toprule
			\multirow{2}{*}{Model} 
			& \multicolumn{4}{c|}{Single-text Obs.} 
			& \multicolumn{8}{c}{Multimodal Obs.} \\ 
			& DA & CV & PAD & EDA & DA & 90\% CI & CV & 90\% CI & PAD & 90\% CI & EDA & 90\% CI \\
			\midrule
			
			Qwen3-VL-30b 
			& 5.42 & 4.07 & 3.97 & 5.27
			& \textbf{5.62 ($\uparrow$3.69\%)} & [0.0, 0.4]
			& \textbf{4.53 ($\uparrow$11.30\%)} & [0.0, 0.9]
			& \textbf{4.17 ($\uparrow$5.04\%)} & [-0.4, 0.7]
			& \textbf{5.58 ($\uparrow$5.88\%)} & [0.0, 0.6] \\ 

			GPT-5-nano 
			& 7.07 & 5.53 & 6.07 & 7.08
			& \textbf{7.32 ($\uparrow$3.54\%)} & [0.1, 0.3]
			& \textbf{5.88 ($\uparrow$6.33\%)} & [-0.3, 0.9]
			& \textbf{6.41 ($\uparrow$5.60\%)} & [0.0, 0.6]
			& \textbf{7.43 ($\uparrow$5.02\%)} & [0.2, 0.5] \\ 

			Grok-4.1-fast 
			& 7.37 & 6.02 & 6.33 & 7.36
			& \textbf{7.52 ($\uparrow$2.04\%)} & [0.0, 0.2]
			& \textbf{6.39 ($\uparrow$6.15\%)} & [0.0, 0.8]
			& \textbf{6.64 ($\uparrow$4.90\%)} & [0.0, 0.7]
			& \textbf{7.57 ($\uparrow$2.72\%)} & [0.0, 0.4] \\ 
			\bottomrule
		\end{tabular}
	} 
\end{table*}

\begin{figure*}[h]
  \centering
  \includegraphics[width=\linewidth]{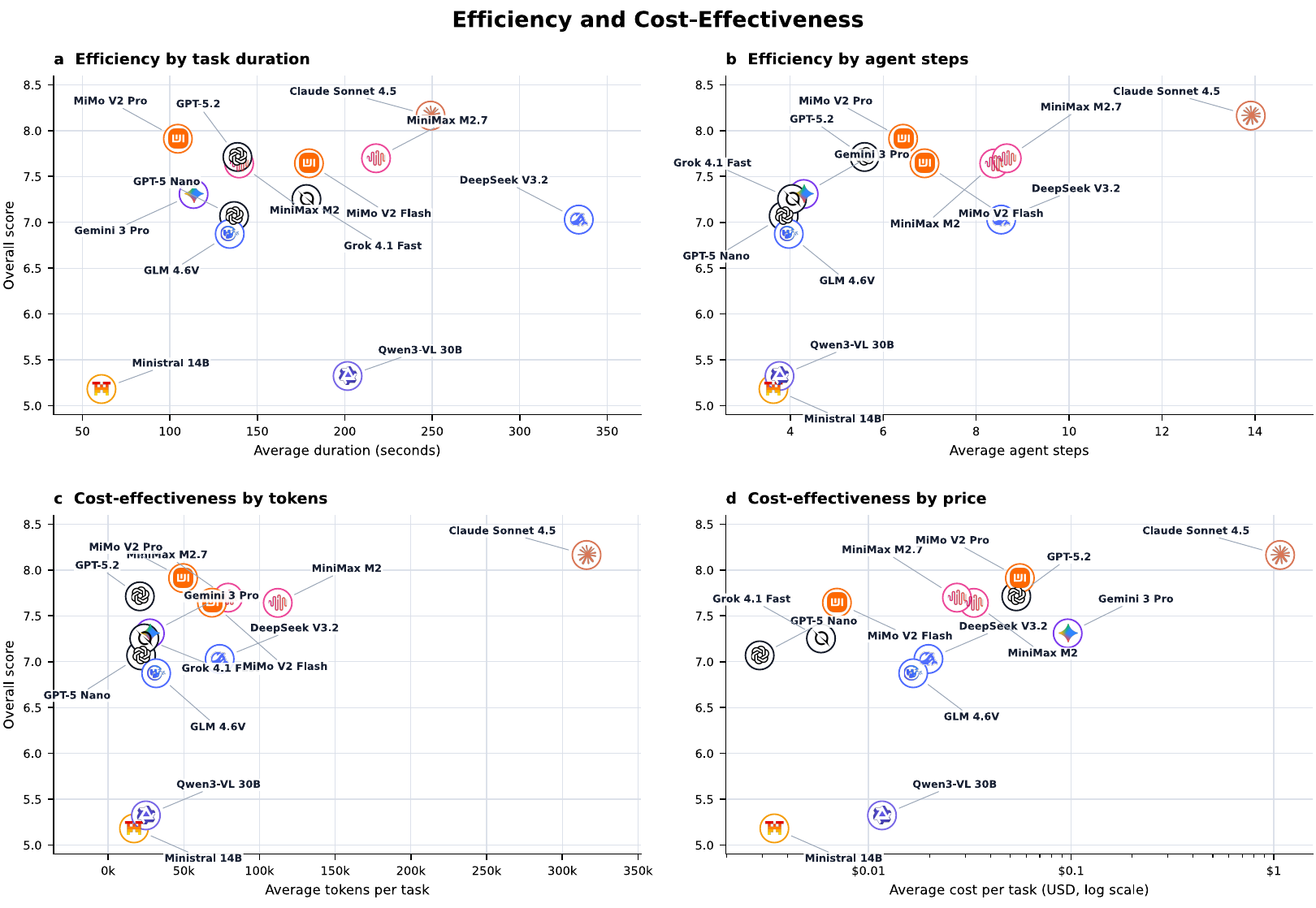}
  \caption{Efficiency and Cost-Effectiveness Analysis. The closer to the top left corner, the better.}
  \label{fig:efficiency}
\end{figure*}

\section{Experiments}
\label{sec:experiments}

In this section, we present a comprehensive empirical evaluation of our proposed framework. Our experiments are designed to address the following three key research questions:

\begin{itemize}[noitemsep]
    \item \textbf{RQ1 (Data Agent Performance):} How do state-of-the-art agents perform on DSAEval, and how does their performance vary across diverse real-world data science domains and complex workflow stages?
    \item \textbf{RQ2 (Efficiency \& Cost-Effectiveness):} What is the trade-off between model performance, operational efficiency (interaction duration of steps), and economic cost?
    \item \textbf{RQ3 (Impact of Multimodal Obs):} Does the ability to perceive multimodal environment observations (e.g., plots) significantly enhance agent performance in vision-related tasks compared to text-only baselines?
\end{itemize}

We further analyze the reliability of our evaluation protocol through aggregation-weight sensitivity, Human--LLM alignment, and inter-judge agreement.

\subsection{Experimental Setup}
For RQ1 and RQ2, we evaluated 13 recent LLMs and VLMs from different organizations, including closed-source models (e.g., GPT-5.2 \citep{openai_gpt52_2025}, Gemini-3-Pro \citep{gemini3pro_modelcard_2025}, Claude-Sonnet-4.5 \citep{claude45sonnet_systemcard_2025}, MiMo-V2-Pro, MiMo-V2-Flash \citep{mimo_v2_flash_api}, GLM-4.6V \citep{glm46v_api}) and open-source models (e.g., MiniMax-M2.7, MiniMax-M2 \citep{minimax_m2_2025},  DeepSeek-V3.2 \citep{deepseek_v32_2025}, Qwen3-VL-30B-A3B-Thinking (hereafter Qwen3-VL-30B) \citep{qwen3technicalreport}, Ministral-3-14B-2512 (hereafter Ministral-3-14B) \citep{ministral3_14b_2512_2025}). These models vary in size, capability, and architectural design, enabling a comprehensive comparison across different model families.
To mitigate potential bias from relying on a single judge model, we initially considered three independent judge models, Claude-Haiku-4.5, GPT-5.1, and Doubao-Seed-1.8, none of which is included in the evaluated agent list.
The final reported scores are based on a validated dual-judge protocol selected through a Human-LLM Alignment Study, which we discuss in Section~\ref{sec:evaluation_reliability}.
Additionally, we set a maximum budget of $T_{\max}=20$ interaction turns per session and a one-hour timeout per iteration to avoid infinite loops, logical deadlocks, and excessive computation. To support computationally intensive tasks such as deep learning training, the sandbox environment was equipped with 4 NVIDIA A100 80GB GPUs. We systematically vary the relative weights assigned to reasoning, code, and final results, and evaluate the stability of the resulting model rankings using Spearman and Kendall rank correlations.

For RQ2, we analyze the task duration, number of agent steps, token consumption, and total cost in USD aggregated over the tasks evaluated in RQ1. We use the model pricing information from OpenRouter\footnote{\url{https://openrouter.ai/models}} to compute the total cost.

For RQ3, the base model must natively support vision modality. We therefore select three eligible models from distinct providers and model families (Qwen3-VL-30B, GPT-5-nano, and Grok-4.1-fast) and use their single-text variants, with $o_t^{\text{img}}$ disabled, as paired baselines. For each model/category combination, we use 10,000 paired bootstrap resamples to estimate a 90\% confidence interval for the mean gap.


\section{Results}
In this section, we present a detailed analysis of the experimental results to answer RQ1, RQ2, and RQ3.
Figure \ref{fig:overall_perf} illustrates the comparative performance of the evaluated models. The results highlight a distinctive hierarchy in agentic data science capabilities.

\paragraph{Overall Performance (RQ1).}
Claude-Sonnet-4.5 emerges as the SoTA model, achieving the highest overall score of 8.164. It outperforms all competing models, including MiMo-V2-Pro (7.912), GPT-5.2 (7.713), Gemini-3-Pro (7.309), and Grok-4.1-fast (7.254). Notably, MiniMax-M2.7 (7.699) closely approaches GPT-5.2 (7.713) and outperforms Gemini-3-Pro (7.309) and Grok-4.1-fast (7.254). In contrast, smaller or more specialized models, including Qwen3-VL-30B (5.324) and Ministral-3-14B (5.182), perform substantially worse on our benchmark. This performance gap indicates that larger models demonstrate a clear performance advantage over smaller models.



To examine fine-grained capabilities, we decompose performance across distinct domains and workflow stages, as depicted in Figure~\ref{fig:radar_analysis}.
\paragraph{Domain Analysis.}
The left chart of Figure \ref{fig:radar_analysis} reveals a clear performance bias towards structured tabular tasks. Models perform exceptionally well in Data Analysis and Business Analytics, where top-tier models approach scores of 8.3. However, performance degrades substantially in unstructured or abstract domains. Specifically, Computer Vision and Natural Language Processing remain the most challenging areas, with all models showing notable performance drops (with average scores of 6.31 and 6.25, respectively). This indicates that while agents have mastered tabular manipulations, they still struggle to handle complex deep learning workflows involving image and text data.

\paragraph{Task Analysis.}
The right chart of Figure \ref{fig:radar_analysis} further highlights the failure modes of current agents. 
Agents exhibit strong performance in the early stages of the data science workflow. In Data Ingestion \& Integration and Data Preparation, top-performing models consistently achieve high scores, indicating effective mastery of standard libraries such as Pandas and NumPy. In contrast, the lowest performance is consistently observed in Prediction \& Forecasting and Model Training \& Optimization, with average scores of 6.03 and 6.49, respectively. Unlike straightforward data cleaning tasks, these stages require iterative experimentation and deeper analytical intuition to identify subtle patterns or effectively tune hyperparameters. This suggests that current agents function more reliably as data analysts than as deep learning engineers, often struggling to handle complex training, prediction, and optimization tasks.


\paragraph{Efficiency \& Cost-Effectiveness (RQ2).}
Figure~\ref{fig:efficiency} summarizes the trade-offs among overall performance, efficiency, and cost.
Notably, although Claude-Sonnet-4.5 achieves the highest overall performance in RQ1, it does so at the cost of substantially higher token consumption, monetary cost, and agent steps.
This suggests that its superior performance is supported by a more resource-intensive reasoning and self-correction process.

For operational efficiency, MiMo-V2-Pro shows the best balance between task duration and performance, while GPT-5.2 achieves strong scores with relatively moderate numbers of agent steps.
For cost-effectiveness, GPT-5.2 is highly token-efficient, whereas MiMo-V2-Flash provides the best monetary value, maintaining competitive performance at an average cost of approximately \$0.007 per task. By contrast, Claude-Sonnet-4.5 incurs the highest cost, approximately \$1.08 per task.

\paragraph{Impact of Multimodal Perception (RQ3).}
Table~\ref{tab:obs_comparison} compares the three image-capable models
against their paired text-only baselines. The results show performance gains for all 12 model–category combinations across DA, CV, PAD, and EDA, ranging from 2.04\% to 11.30\%. Qwen3-VL-30b has the largest aggregate gain (11.30\%) on CV. In representative trajectories, visual feedback changes subsequent actions: Qwen reads word-frequency plots to ground an Animal Crossing review comparison, while Grok revises contour thresholds and digit spacing after inspecting MNIST diagnostics.

\subsection{Evaluation Reliability Analysis}
\label{sec:evaluation_reliability}

\paragraph{Weight Sensitivity.}
Table~\ref{tab:weight_sensitivity} compares the agent rankings obtained under alternative weights for
$S_{\text{reasoning}}$, $S_{\text{code}}$, and $S_{\text{result}}$
against the default ranking. The rankings remain highly stable across all representative settings, and the top-ranked model changes only under the code-only configuration.

\begin{table}[h]
\centering
\footnotesize
\setlength{\tabcolsep}{2pt}
\begin{tabularx}{\columnwidth}{@{}>{\raggedright\arraybackslash}Xcc@{}}
\toprule
\textbf{Weights (R/C/F)} & \textbf{Spearman} & \textbf{Kendall $\tau_b$} \\
\midrule
Equal (1/3, 1/3, 1/3) & 1.000 & 1.000 \\
Reason-heavy (.50/.25/.25) & 1.000 & 1.000 \\
Code-heavy (.25/.50/.25) & 0.995 & 0.974 \\
Result-heavy (.25/.25/.50) & 0.995 & 0.974 \\
Reason-only (1/0/0) & 0.995 & 0.974 \\
Code-only (0/1/0) & 0.940 & 0.846 \\
Result-only (0/0/1) & 0.956 & 0.872 \\
\bottomrule
\end{tabularx}
\caption{Sensitivity of agent rankings to representative aggregation weights. Correlations compare each 13-agent ranking with the default ranking.}
\label{tab:weight_sensitivity}
\end{table}

\paragraph{Human--LLM Calibration.}
Three Ph.D. candidates in data science annotated the complete 641-task log of Gemini-3-Pro, which was randomly selected as the calibration agent, and their consensus was used as the human reference. As shown in Table~\ref{tab:judge_alignment_mse}, GPT-5.1 and Claude-Haiku-4.5 achieve substantially lower overall MSEs (0.855 and 0.896) than Doubao-Seed-1.8 (4.530), which we therefore exclude. Accordingly, the final score is the task-level mean of the two retained judges (Claude-Haiku-4.5 and GPT-5.1).

\begin{table}[h]
\centering
\small
\setlength{\tabcolsep}{4pt}
\renewcommand{\arraystretch}{1.12}
\begin{tabular}{lcccc}
\toprule
\textbf{Judge Model} & \textbf{Reason} & \textbf{Code} & \textbf{Result} & \cellcolor{gray!15}\textbf{Overall} \\
\midrule
GPT-5.1          & 1.143 & 1.181 & 1.857 & \cellcolor{gray!15}0.855 \\
Claude-Haiku-4.5 & 1.038 & 0.975 & 2.031 & \cellcolor{gray!15}0.896 \\
Doubao-Seed-1.8  & 3.225 & 3.837 & 8.913 & \cellcolor{gray!15}4.530 \\
\bottomrule
\end{tabular}
\caption{Human--LLM alignment measured by MSE between each candidate judge and human consensus. Lower is better.}
\label{tab:judge_alignment_mse}
\end{table}

\paragraph{Human--Judge Agreement.}
We further investigated the agreement between the human consensus and the dual-judge mean. We measure score-level agreement using Pearson correlation and MAE, and rank the outputs by their scores to calculate Spearman correlation and Kendall's $\tau_b$. From Table~\ref{tab:human_dual_agreement}, the dual-judge mean strongly agrees with human consensus in overall scores (Pearson $=0.872$; MAE $=0.364$) and task ordering (Spearman $=0.895$; Kendall's $\tau_b=0.773$).

\begin{table}[H]
\centering
\scriptsize
\renewcommand{\arraystretch}{1.06}
\begin{tabular*}{\columnwidth}{@{\extracolsep{\fill}}llcc@{}}
\toprule
\textbf{Analysis level} & \textbf{Dimension} & \textbf{Pearson} & \textbf{MAE} \\
\midrule
\multirow{4}{*}{\shortstack[l]{Score\\level}}
& Reasoning    & 0.770 & 0.471 \\
& Code         & 0.764 & 0.516 \\
& Final result & 0.887 & 0.570 \\
& Overall      & \textbf{0.872} & \textbf{0.364} \\
\midrule
\textbf{Analysis level} & \textbf{Dimension} & \textbf{Spearman} & \textbf{Kendall $\tau_b$} \\
\midrule
\multirow{4}{*}{\shortstack[l]{Ranking\\level}}
& Reasoning    & 0.792 & 0.673 \\
& Code         & 0.744 & 0.623 \\
& Final result & 0.892 & 0.787 \\
& Overall      & \textbf{0.895} & \textbf{0.773} \\
\bottomrule
\end{tabular*}
\caption{Score-level and ranking-level agreement between human consensus and the final dual-judge mean.}
\label{tab:human_dual_agreement}
\end{table}

\paragraph{Inter-Judge Agreement.}
We compare the scores assigned by the two judges to 8,314 valid model task outputs from all 13 agents. As shown in Table~\ref{tab:judge_agreement}, their task-level overall scores achieve a Pearson correlation of 0.813 with an MAE of 1.004. After averaging the task scores by agent, the resulting rankings achieve a Spearman correlation of 0.945 and Kendall's $\tau_b$ of 0.846. Domain-level results are provided in Appendix~\ref{app:judge_agreement}.

\begin{table}[H]
\centering
\scriptsize
\renewcommand{\arraystretch}{1.06}
\begin{tabular*}{\columnwidth}{@{\extracolsep{\fill}}llcc@{}}
\toprule
\textbf{Analysis level} & \textbf{Dimension} & \textbf{Pearson} & \textbf{MAE} \\
\midrule
\multirow{4}{*}{\shortstack[l]{Score\\level}}
& Reasoning    & 0.747 & 1.043 \\
& Code         & 0.855 & 0.739 \\
& Final result & 0.711 & 1.414 \\
& Overall      & \textbf{0.813} & \textbf{1.004} \\
\midrule
\textbf{Analysis level} & \textbf{Dimension} & \textbf{Spearman} & \textbf{Kendall $\tau_b$} \\
\midrule
\multirow{4}{*}{\shortstack[l]{Ranking\\level}}
& Reasoning    & 0.984 & 0.923 \\
& Code         & 0.978 & 0.923 \\
& Final result & 0.907 & 0.769 \\
& Overall      & \textbf{0.945} & \textbf{0.846} \\
\bottomrule
\end{tabular*}
\caption{Agreement between GPT-5.1 and Claude-Haiku-4.5 across all evaluated agents.}
\label{tab:judge_agreement}
\end{table}

\section{Future Directions}

Our results indicate that future data science agents should place greater emphasis on robust, execution-grounded workflow completion.
Current models still struggle in unstructured domains such as Computer Vision and NLP, where tasks require complex data loading, custom preprocessing, model architecture design, and iterative debugging. The quantified audit in Appendix~\ref{sec:qualitative_error_analysis} shows that incomplete requirement coverage (52.1\%) and visualization/result interpretation (41.8\%) are the most prevalent capability-gap indicators among imperfect trajectories. Future systems should therefore strengthen intermediate-state verification and enforce consistency between execution traces, generated artifacts, and final conclusions.


Furthermore, DSAEval produces real experimental traces from authentic datasets, including multi-step interactions, execution outputs, failure cases, and final reports. These traces can serve as valuable resources for training, debugging, and evaluating data agents in reinforcement learning and self-improvement settings.
A dedicated task viewer\footnote{\url{https://dsaeval.lambda.org.ai/}} provides access to model results and detailed task logs.
Future benchmark extensions should add more image-capable model families, stratified human evaluation across agent styles and domains, more unstructured tasks, and hidden or dynamically swapped datasets for stronger contamination control.





\section{Conclusion}
\label{sec:conclusion}
We introduce DSAEval, a benchmark for evaluating autonomous data science agents on real-world workflows. It supports multimodal environment perception, multi-query interactions, and multi-dimensional evaluation across reasoning, code, and final results. By evaluating 13 recent LLMs and VLMs, we find that Claude-Sonnet-4.5 achieves the strongest overall performance, while MiMo-V2-Pro, GPT-5.2, and MiMo-V2-Flash lead in duration efficiency, token/step efficiency, and cost-effectiveness, respectively. Multimodal perception improves vision-related tasks by up to 11.30\%. Overall, current agents handle structured data and routine workflows well, but still struggle with unstructured domains and complex modeling tasks. DSAEval provides a rigorous testbed for developing next-generation data science agents.

\section*{Limitations}
Despite the rigorous design of DSAEval, we acknowledge specific limitations inherent to our methodology.

\paragraph{Randomness and Human Subjectivity}
LLM randomness may produce variations across repeated agent runs and judge evaluations, while human assessment of open-ended reasoning, code, and reports may also be subjective. We mitigate these effects through a fixed evaluation protocol, consistent scoring dimensions, and aggregated human consensus, although some variation remains unavoidable.

\paragraph{Data Leakage and Contamination}
Because some source datasets and notebooks are publicly available on platforms such as Kaggle, models may have encountered related data or solutions during pre-training. New QRA formulations and dynamic sandbox execution reduce direct memorization but cannot eliminate this risk. Future versions will incorporate newly collected datasets and dynamic dataset swapping to strengthen contamination control.

\paragraph{Scope and Benchmark Balance}
DSAEval evaluates problem solving with provided datasets and task contexts, but does not cover requirements elicitation, data governance, or production deployment. Moreover, 79.1\% of the data instances are tabular and 54.3\% of the tasks are labeled as Data Analysis. Although this reflects common data-science practice, the aggregate results should not be interpreted as uniformly representing all data modalities and domains.

\paragraph{Judge Validation Scope}
Human validation covers all 641 Gemini-3-Pro trajectories but not all evaluated agents. Although the two judges show strong cross-model agreement, broader human validation across agents and domains remains necessary. The multimodal ablation is also limited to three image-capable model families, and future work should evaluate more multimodal agents.

\section*{Ethical Considerations}
All agent-generated code was executed in an isolated sandbox to limit risks to external data, programs, and infrastructure. DSAEval uses publicly available datasets, which we reviewed for personally identifiable information and used in accordance with their licenses and terms of use. Human annotators participated voluntarily, were informed of the study purpose, and provided informed consent before annotation. As the study collected no personal or sensitive data and involved only evaluation of model outputs, formal IRB review was not required under our institutional policy.



\section*{Acknowledgments}
This work was supported by The Hong Kong Polytechnic University through Research Grant P0046811. The research of Ruijian Han was supported by the Hong Kong Research Grants Council General Research Fund (No. 15302225) and by research grants from The Hong Kong Polytechnic University (P0050935, P0063848). The research of Binyan Jiang was partially supported by the Hong Kong Research Grants Council General Research Fund (15302924). The research of Defeng Sun and Yancheng Yuan was partially supported by the Research Center for Intelligent Operations Research at The Hong Kong Polytechnic University (P0051214).

During the preparation of this work, the authors used ChatGPT and Gemini for language editing and polishing. All AI-assisted content  was reviewed by the authors.


\bibliography{custom.bib}

\appendix
\clearpage
\onecolumn

\section{Data Sources}\label{append_data}
 \begin{table}[H]
  \centering
  \caption{Kaggle datasets by Domains}
  \label{tab:kaggle-sources}
  \begin{tabularx}{\textwidth}{l X}
   \toprule
   \textbf{Domains} & \textbf{Kaggle Ipynb Name} \\
   \midrule
   \textbf{Data Analysis} & fifa19-data-cleaning.ipynb \\
   & indian-grocery-supermarket-big-basket-eda.ipynb \\
   & exploring-spotify-data.ipynb \\
   & $\cdots$ \\
  
   \addlinespace[0.5em] 
   \textbf{Time Series} & ocean-wave-prediction-with-lstm.ipynb \\
   & easy-wind-power-forecasting.ipynb \\
   & us-border-crossing-eda-and-forecasting.ipynb \\
   & $\cdots$ \\
   \addlinespace[0.5em]
   \textbf{Domain-Specific Applications} & learnt-parameters-for-under-3-5-goals.ipynb \\
   & formula-1-post-race-summary.ipynb \\
   & rna-seq-salmon-tximport-pipeline-vol-1.ipynb \\
   & $\cdots$ \\
   \addlinespace[0.5em]
   \textbf{Natural Language Processing} & classify-emotions-in-text-with-bert.ipynb \\
   & social-media-analysis-sentiment.ipynb \\
   &songs-similarities-by-lyrics-scraping-and-analysis.ipynb\\
   & $\cdots$ \\
   \addlinespace[0.5em] 
   \textbf{Computer Vision} & shoe-vs-sandal-vs-boot-multiclass-acc-0-98.ipynb \\
   & hand-sign-multi-class-classification-cnn-97.ipynb \\
   & brain-tumor-segmentation-detectron2-map-50-76-2.ipynb \\
   & $\cdots$ \\
   \addlinespace[0.5em]
   \textbf{Business Analytics} & predict-product-price-sephora-website-rmse-0-078.ipynb \\
   & data-analysis-for-marketing-strategy.ipynb \\
   & market-basket-analysis-with-apriori.ipynb\\
   & $\cdots$ \\
    \addlinespace[0.5em] 
   \textbf{Clustering} & clustering-practice-k-means-analysisfor-beginners.ipynb \\
   & shop-customer-clustering.ipynb \\
   & classify-gamers-mentality.ipynb \\
   & $\cdots$ \\
   \addlinespace[0.5em]
   \textbf{Statistical Testing \& Experimentation} 
    & cern-electron-collision-prediction.ipynb\\
    & amazon-s-books-eda-plotly-hypothesis-test.ipynb \\
   & survival-analysis-with-cox-model-implementation.ipynb \\
  
   & $\cdots$ \\
   \addlinespace[0.5em]
   \textbf{Others} & starter-tufts-face-database-dbb85c33-d.ipynb \\
   & exercise-syntax-variables-and-numbers.ipynb \\
   & how-to-retrieve-gcs-paths-from-kaggle-datasets.ipynb\\
   & $\cdots$ \\
   
   \bottomrule
  \end{tabularx}
 \end{table}

 \begin{table}[H]
 \centering
 \caption{Textbook datasets by Domain}
 \label{tab:textbook-sources}
 \begin{tabularx}{\textwidth}{l X}
  \toprule
  \textbf{Domains} & \textbf{Textbook Name} \\
  \midrule
  \textbf{Machine Learning} & Applied Machine Learning with Python\\
  & Machine Learning with PyTorch and Scikit-Learn \\
  & Deep Learning for Time Series Forecasting \\
  & $\cdots$ \\
  
  \addlinespace[0.5em] 
  \textbf{Statistical Learning} & An Introduction to Statistical Learning \\
  & Regression Modeling Strategies \\
  & Survival Analysis Using S: Analysis of Time-to-Event Data \\
  & $\cdots$ \\
  \addlinespace[0.5em]
  \textbf{Data Analysis} & ggplot2: Elegant Graphics for Data Analysis \\
  &Data Engineering with Python \\
  & Data Visualization: A Practical Introduction \\
  & $\cdots$ \\
  \bottomrule
 \end{tabularx}
\end{table}

\begin{multicols}{2}
\raggedcolumns
 
  

\section{Benchmark Data Validation and Quality Control}
\label{app:data_validation}

To ensure the quality and reliability of DSAEval, we conducted a multi-stage quality control process combining programmatic checks and expert human validation.
We began with a large pool of 2,421 synthesized candidate QRA pairs.
Each candidate pair contains a question, a reference reasoning process, and a reference answer generated from real data science cases.

\paragraph{Expert Annotation.}
We employed two expert annotators to manually review the synthesized QRA pairs.
The validation was based on three strict rejection criteria:

\begin{itemize}
    \item \textbf{Data Accessibility.}
    The corresponding dataset must be accessible, valid, and loadable in the sandbox environment.
    Candidates with missing files, invalid paths, corrupted data, or data loading failures were rejected.

    \item \textbf{Question Clarity.}
    The question must be unambiguous and solvable based on the provided dataset and task context.
    Candidates were rejected if the problem statement was underspecified, overly vague, inconsistent with the dataset, or dependent on information unavailable to the agent.

    \item \textbf{Answer Validity.}
    The reference reasoning process and answer serve as soft ground truth.
    Therefore, the reasoning must be logically sound, free from major methodological errors, and directly relevant to the question.
    Candidates were rejected if the reference reasoning contained clear logical mistakes, if the answer did not address the question, or if the code-generated result was unreliable.
\end{itemize}

\paragraph{Filtering Results.}
Through programmatic quality checks and expert validation, approximately 73.5\% of the synthesized candidates were rejected.
Specifically, 1,780 out of 2,421 candidate QRA pairs were filtered out due to issues such as data loading failures, ambiguous problem descriptions, invalid reference answers, or low-quality code generation.
The final benchmark contains 641 problems, representing the top 26.5\% high-quality subset that passed this strict multi-stage filtering process.
This filtering procedure helps ensure that DSAEval focuses on valid, executable, and clearly specified real-world data science tasks.

\subsection{Representative Accepted and Rejected QRA Pairs}

\paragraph{Accepted QRA: Animal Crossing review analysis (ID 92).}
\textbf{Question:} Starting from the Animal Crossing \texttt{user\_reviews.csv} dataset, create cleaned word-frequency distributions for Negative and Positive review corpora and use Shifterator to construct an entropy-shift graph quantifying which words drive the differences between the two sets. Provide the resulting visualization and summarize what the graph encodes.

\textbf{Reasoning:} Load the dataset and split the reviews into Negative and Positive groups using the median grade as the boundary. Clean both corpora by removing punctuation, common stop words, and domain-specific tokens, and convert text to lowercase. Count word frequencies for each group to obtain comparable distributions. Use an entropy-based word-shift approach to contrast the distributions and quantify each word's contribution to the difference. Render the shift graph to visualize which words increase or decrease entropy between Negative (reference) and Positive (comparison) review texts, with colors indicating which corpus a word is associated with.

\textbf{Answer:} An entropy-shift graph comparing Negative (reference) and Positive (comparison) reviews was produced. The visualization shows word-level contributions to the difference in usage, with Negative-associated contributions displayed in purple and Positive-associated contributions in yellow. Specific top-contributing words are visible on the chart. \texttt{<image\_id:4>}

\textbf{Why accepted:} The task is fully supported by the available raw review data and has a coherent end-to-end boundary. The reasoning matches the executed cleaning, frequency-counting, and entropy-shift workflow, while the answer and color interpretation are grounded in the saved visualization. It therefore satisfies all three validation criteria: data accessibility, question clarity and scope, and reasoning and answer validity.

\paragraph{Rejected QRA: COVID-19 Prophet forecasting (ID 6325).}
\textbf{Question:} From the aggregated global daily cases series, fit a Prophet model with weekly seasonality only, forecast the next 30 days at daily frequency, report the in-sample $R^2$, and visualize the forecast with prediction intervals.

\textbf{Reasoning:} First, build a two-column time series with dates and global daily case counts and convert the date column to a proper datetime index. Initialize a forecasting model configured for weekly seasonality to capture weekly patterns, and disable daily and yearly seasonality to match the chosen specification. Fit the model on the historical data, generate a 30-day future dataframe at daily resolution, and produce point forecasts with upper and lower uncertainty bounds. Evaluate the model's fit by comparing predicted values against actuals over the training period using $R^2$. Finally, visualize the forecasted trajectory with shaded intervals to convey uncertainty.

\textbf{Answer:} The 30-day daily forecast was generated with Prophet (weekly seasonality enabled), and the forecast plot with prediction intervals was produced. The in-sample $R^2$ is \texttt{[uncertain]}. \texttt{<image\_id:2>}

\textbf{Why rejected:} Although the dataset and forecasting procedure are relevant, the question explicitly requires a numerical in-sample $R^2$, while the saved notebook contains no retained numerical output for that call. Consequently, the synthesized answer leaves a required result as \texttt{[uncertain]}, making the reference incomplete and not independently verifiable. The candidate was therefore rejected on reference-answer validity and execution grounding. This example illustrates how human validation removes unsupported or incomplete references even when the underlying task itself appears reasonable.

\end{multicols}
\section{Detail Prompt}
\label{sec:detail_prompt}

\begin{promptbox}{Prompt for Generating QRA Pairs}{cyan}
	\ttfamily\small 
	\indent You are the examiner, and you need to create some data science questions based on a answered jupyter notebook for exam. \\
	\\
	\indent Your mission is to carefully analyze the given Jupyter Notebook and generate a list of high-quality data science 'Question–Reasoning–Answer' (QRA) triplets. These triplets will serve as benchmark data to evaluate LLM performance on data science tasks. \\
	\\
	\indent ==================== \\
	\indent \#\#\# Important Requirements \\
	\\
	\indent - The task granularity is flexible. For example, a "modeling" task may encompass a full workflow (data loading $\to$ cleaning $\to$ visualization $\to$ feature engineering $\to$ modeling $\to$ evaluation) if these steps form a coherent chain. \\
	\indent - However, every reasoning chain must start from the raw data or a logically prior step, not from a later checkpoint (e.g., do NOT start with “load the pre-trained model A1” unless the notebook itself explicitly includes that step). \\
	\\

	\indent \#\#\# Output Format \\
	\\
	\indent You MUST return a valid JSON object in the following structure:\\
	\indent [ \\
	\indent \quad \{ \\
	\indent \quad \quad "data\_type" : "one or more of the data type used in the task...", \\
	\indent \quad \quad "domain" : "one or more of the notebook's task domain...", \\
	\indent \quad \quad "task\_type": "one or more of the data science task categories...", \\
	\indent \quad \quad "language": "The programming languages, (e.g. Python)", \\
	\indent \quad \quad "question": "A clear, specific data science question...", \\
	\indent \quad \quad "reasoning": "A detailed, step-by-step logical explanation...", \\
	\indent \quad \quad "answer": "The specific answer... For data visualization, keeping the image id in the last (e.g. <image\_id:1>).", \\
	\indent \quad \quad "best\_score (Optional)": "This is only valid for modeling and evaluation...", \\
	\indent \quad \quad "confidence": "The confidence score for the answer... (range from 1 to 4)" \\
	\indent \quad \}, \\
	\indent \quad ... \\
	\indent ]
	
	\indent \#\#\# Detailed Instructions \\
	\indent 1. Data type and domain \\
	\indent \quad - Typically, the data type and domain of QRA are the same in each notebook. \\
	\\
	\indent 2. Question \\
	\indent \quad - Must be a data science questions in user style with data and target (e.g. I have the data mlg-ulb/creditcardfraud...). \\
	\indent \quad - Should align with a clear analytical step such as dataset inspection, preprocessing, visualization... \\
	\indent \quad - DO not ask question directly within the notebook. \\
	\indent \quad - MUST start from row data, can not ask questions depend on the last step. \\
	\\
	\indent 3. Reasoning \\
	\indent \quad - Must provide a comprehensive, logically connected, and step-by-step explanation... \\
	\indent \quad - Do not include or describe code syntax or variable names. \\
	\indent \quad - The reasoning should form a clear “chain of thought” from raw data understanding to final result. \\
	\\
	\indent 4. Answer \\
	\indent \quad - The answer MUST be real output from the notebook. \\
	\indent \quad - For data visualization, keeping the image id in the last (e.g. <image\_id:1>). \\
	\indent \quad - For the confidence score. 1 is totally uncertain, 4 is totally certain. \\
	\\
	\indent 5. Quality Requirements \\
	\indent \quad - Do no mention the notebook hence the student can not see a notebook. \\
	\indent \quad - Ensure diversity across task types. \\
	\indent \quad - The final output must be valid JSON, parsable without syntax errors. \\
	\\
	\indent ==================== \\
	\indent \#\#\# Example \\
	\indent [ \\
	\indent \quad \{ \\
	\indent \quad \quad "data\_type" : "tabular data", "domain" : "exploratory data analysis", \\
	\indent \quad \quad "task\_type": "data visualization", "language": "Python", \\
	\indent \quad \quad "question": "Given the breast cancer survival dataset... what is the distribution...?", \\
	\indent \quad \quad "reasoning": "First, the dataset must be loaded... Then... Lastly, use matplotlib...", \\
	\indent \quad \quad "answer": "...The Tumor Stage distribution shows... <image\_id:1>", \\
	\indent \quad \quad "confidence": "3" \\
	\indent \quad \}, ... \\
	\indent ] \\
	
	\vspace{1.5em}
	\indent Here is the Jupyter Notebook content, ================== is the separator for each cell or output: \\
	\\
	\indent [START OF NOTEBOOK CONTENT] \\
	\\
	\indent \textbf{\{notebook\_content\}} \\
	\\
	\indent [END OF NOTEBOOK CONTENT] \\
	\\
	\indent Now, generating \textbf{\{qra\_numbers\}} long QRA triplets based on this content. \\
\end{promptbox}

\begin{promptbox}{Data Science Agent Action Prompt (VLM)}{cyan}
 \ttfamily\small 
 \indent You are an autonomous data and code execution agent running in a sandbox environment who helps the user to do data related tasks. \\
 \\
 Environment and capabilities: \\
 \indent - For the user's task, you should write Python code step by step and execute it using the Jupyter Notebook tools, and you will see the execution results. \\
 \indent - For complex problem, you'd better to solve the task by multi-step, you can reuse the defined variable before. \\
 \indent - Commonly used data-science-related packages are already prepared in the environment. But if a package is missing, you can run \texttt{!pip install <package>} to install it in the Jupyter kernel. \\
 \indent - The user's data files are located at: \textbf{\{data\_path\}}. \\
 \indent - There are 4 * NVIDIA A100-SXM4-80GB GPUs to use. However, to conserve system resources, the GPU should ONLY be used when necessary. You cannot stop other programs that are using the GPU. Besides, under no circumstances should any actions modify CUDA-related versions or install packages that could affect the existing CUDA environment. To conserve system resources, the GPU should only be used when strictly necessary. \\
 \indent Under no circumstances should any actions modify CUDA-related versions or install packages that could affect the existing CUDA environment, \\
 \indent and you MUST not terminate or clear any GPU processes, as other programs may be using them. \\
 \indent - IMPORTANTLY, you can only have a maximum of 20 steps/iteration opportunities (from 0-19). Your session will be forcibly terminated after the limit is exceeded. Besides, the longest code execution time in the kernel is 1 hour, so you should choose the option that can finish running as quickly as possible and use less steps. \\
 \indent - You may save temporary or auxiliary outputs (e.g., intermediate caches, figures, logs) to: \textbf{\{working\_path\}}. \\
 \indent - You have vision ability, you can see plotted figure by '\%matplotlib inline' in you code. \\
 \indent - Finally, you MUST produce a final report based on your experiments and findings in markdown format. \\
 \\
 General Instructions: \\
 \indent - Load data only from \textbf{\{data\_path\}} and save outputs only in \textbf{\{working\_path\}}. \\
 \indent - If multiple files are available, identify which ones are relevant and briefly justify your choice in the final explanation. \\
 \indent - If you need to plot figures, you MUST add '\%matplotlib inline' in your code. \\
 \indent - You can not stop other programs that are using the GPU and take any actions that may modify or harm the CUDA environment. Besides, for applications that require GPUs, you should try to use PyTorch instead of TensorFlow. \\
 \indent - You should aim for reproducible results, such as setting a random seed in applicable tasks. \\
 \indent - Try your best to produce a complete and correct solution. \\
 \\
 *Final answer format* (strictly follow this structure): \\
 \\
 \#\# Final Output \\
 \\
 1. Task Understanding \\
 \indent Briefly restate the problem and the main goal in 2--5 sentences. \\
 \\
 2. Approach Summary \\
 \indent Describe your solution strategy in clear, high-level steps (conceptual only, not low-level inner thoughts). \\
 \indent Explain how you use the data, which files you select, and any important modeling or analysis choices. \\
 \\
 3. Key Implementation \\
 \indent Provide some piece of key python codes of your problem solving process with explanations. \\
 \\
 4. Results \& Explanation \\
 \indent - Report the key numeric results, metrics, or findings explicitly (e.g., final metrics, key statistics). \\
 \indent - Summarize what these results mean in a concise, well-structured text. \\
 \indent - If you generated any important artifacts (e.g., figures, model files), mention their filenames and what they contain. \\
 \indent - Do NOT provide or reference any separate report file; all essential information must be included in this section. \\
 \\
 Important constraints: \\
 \indent - Always add '\%matplotlib inline' in your plotting code. \\
 \indent - Return ONLY the final output in the structure above. \\
 \indent - Do not omit numeric results; they must appear directly in Results \& Explanation. \\
 """
\end{promptbox}

\begin{promptbox}{Data Science Agent Action Prompt (LLM)}{cyan}
 \ttfamily\small 
 \indent You are an autonomous data and code execution agent running in a sandbox environment who helps the user to do data related tasks. \\
 \\
 Environment and capabilities: \\
 \indent - For the user's task, you should write Python code step by step and execute it using the Jupyter Notebook tools, and you will see the execution results. \\
 \indent - For complex problem, you'd better to solve the task by multi-step, you can reuse the defined variable before. \\
 \indent - Commonly used data-science-related packages are already prepared in the environment. But if a package is missing, you can run \texttt{!pip install <package>} to install it in the Jupyter kernel. \\
 \indent - There are 4 * NVIDIA A100-SXM4-80GB GPUs to use. However, to conserve system resources, the GPU should ONLY be used when necessary. Under no circumstances should any actions modify CUDA-related versions or install packages that could affect the existing CUDA environment, To conserve system resources, the GPU should only be used when strictly necessary. \\
 \indent Under no circumstances should any actions modify CUDA-related versions or install packages that could affect the existing CUDA environment, \\
 \indent and you MUST not terminate or clear any GPU processes, as other programs may be using them. \\
 \indent - IMPORTANTLY, you can only have a maximum of 20 steps/iteration opportunities (from 0-19). Your session will be forcibly terminated after the limit is exceeded. Besides, the longest code execution time in the kernel is 1 hour, so you should choose the option that can finish running as quickly as possible. \\
 \indent - The user's data files are located at: \textbf{\{data\_path\}}. \\
 \indent - You may save temporary or auxiliary outputs (e.g., intermediate caches, figures, logs) to: \textbf{\{working\_path\}}. \\
 \indent - You do not have vision ability. So, you can save figures directly using \texttt{plt.savefig()}. \\
 \indent - Finally, you MUST produce a final report based on your experiments and findings in markdown format. \\
 \\
 General Instructions: \\
 \indent - Load data only from \textbf{\{data\_path\}} and save outputs only in \textbf{\{working\_path\}}. \\
 \indent - If multiple files are available, identify which ones are relevant and briefly justify your choice in the final explanation. \\
 \indent - Save all figures and models or other files you think valuable. \\
 \indent - You can not stop other programs that are using the GPU and take any actions that may modify or harm the CUDA environment. Besides, for applications that require GPUs, you should try to use PyTorch instead of TensorFlow. \\
 \indent - You should aim for reproducible results, such as setting a random seed in applicable tasks. \\
 \indent - Try your best to produce a complete and correct solution. \\
 \\
 *Final answer format* (strictly follow this structure): \\
 \\
 \#\# Final Output \\
 \\
 1. Task Understanding \\
 \indent Briefly restate the problem and the main goal in 2--5 sentences. \\
 \\
 2. Approach Summary \\
 \indent Describe your solution strategy in clear, high-level steps (conceptual only, not low-level inner thoughts). \\
 \indent Explain how you use the data, which files you select, and any important modeling or analysis choices. \\
 \\
 3. Key Implementation \\
 \indent Provide some piece of key python codes of your problem solving process with explanations. \\
 \\
 4. Results \& Explanation \\
 \indent - Report the key numeric results, metrics, or findings explicitly (e.g., final metrics, key statistics). \\
 \indent - Summarize what these results mean in a concise, well-structured text. \\
 \indent - If you generated any important artifacts (e.g., figures, model files), mention their filenames and what they contain. \\
 \indent - Do NOT provide or reference any separate report file; all essential information must be included in this section. \\
 \\
 Important constraints: \\
 \indent - Return ONLY the final output in the structure above. \\
 \indent - Do not omit numeric results; they must appear directly in Results \& Explanation. \\
 """
\end{promptbox}

\begin{promptbox}{Multimodal Evaluation Prompt for Data Science Tasks}{cyan} \label{app:judeg_pmt}
	\ttfamily\small
	You are a data science evaluation assistant. Your task is to objectively evaluate the quality of a generated solution for the given \textbf{[PROBLEM]}. \\
	\\
	\#\#\# Evaluation Principles: \\
	1. Core Objective: Did the model answer the \textbf{[PROBLEM]} correctly? \\
	2. The \textbf{[STANDARD\_ANSWER]} represents just one possible way. The \textbf{[predicted\_reasoning\_answer]} may adopt a different but equally valid approach... \\
	3. Visuals: You may receive "Reference Figures" and "Predicted Figures". \\
	\indent - DO NOT PENALIZE if the predicted answer has NO figures, *UNLESS* the \textbf{[PROBLEM]} explicitly asks to "plot", "visualize", or "graph". \\
	\indent - If both have figures, compare them to ensure the predicted figure shows correct data distribution, trends and insights. \\
	4. Code-Answer Consistency: You must verify if the final answer in \textbf{[predicted\_reasoning\_answer]} is actually derived from \textbf{[PREDICTED\_CODE]}. \\
	\\
	\#\#\# Evaluation Dimensions: \\
	\\
	* Consistency (true or false): Focus on the alignment between \textbf{[PREDICTED\_CODE]} and \textbf{[predicted\_reasoning\_answer]}. \\
	\indent - true: The narrative accurately reflects the code's final output. \\
	\indent - false: The narrative claims results that the code did not produce. \\
	\\
	* ReasoningProcess (Score 0-10): Focus on the conceptual soundness and validity of the logic. \\
	\indent - 0-2 (Poor): Fundamental errors or violates core statistical principles. \\
	\indent - 5-6 (Fair): Partially aligns but has obvious logical defects. \\
	\indent - 9-10 (Excellent): Conceptually clear and logically fully valid. \\
	\\
	* CodeSteps (Score 0-10): Focus on the correctness and completeness of the \textbf{[PREDICTED\_CODE]}. \\
	\indent - 0-2 (Poor): Codes are missing, disordered, or mathematically incorrect. \\
	\indent - 9-10 (Excellent): All major steps are correct, coherent, and lead logically toward the result. \\
	\\
	* FinalResults General Principles: \\
	\indent (1) Quantitative tasks (e.g., MSE, Accuracy): Judge based on whether the reported value is reasonable. \\
	\indent (2) Qualitative tasks: Judge whether the conclusion logically follows from prior reasoning. \\
	\indent (3) Visualizations: Evaluate Alignment, Effectiveness, Readability, and Aesthetics. \\
	\\
	\#\#\# Holistic Scoring (Integrating Text and Visuals): \\
	- Score 5-6 (Fair - Baseline): The result is correct and achieves the same level of quality as the standard answer. \\
	- Score 9-10 (Excellent): The result is objectively correct and significantly superior to the standard answer (e.g., publication-quality visuals). \\
	\\
	\textbf{[PROBLEM]:} \{problem\} \\
	\\
	\textbf{[STANDARD\_ANSWER]:} \\
	\{standard\_answer\} \\
	\\
	\textbf{[PREDICTED\_CODE]:} \\
	\{predicted\_code\} \\
	\\
	\textbf{[predicted\_reasoning\_answer]:} \\
	\{predicted\_reasoning\_answer\} \\
	\\
	\# Output Format \\
	Return only the following JSON, without any additional text or code:
	
	\indent [ \\
	\indent \quad \{ \\
	\indent \quad \quad "Analysis": "A brief explanation of why these scores were given", \\
	\indent \quad \quad "ReasoningProcess": <score>, \\
	\indent \quad \quad "CodeSteps": <score>, \\
	\indent \quad \quad "FinalResults": <score>, \\
	\indent \quad \quad "Consistency": <true or false> \\
	\indent \quad \} \\
	\indent ]
	
\end{promptbox}

\section{Generated QRA Examples}
\label{sec:Generated QRA Examples}
\begin{promptbox}{QRA Example 1}{blue}
	\ttfamily\small 
	\indent "data\_type": "text data", \\
	\indent "domain": "Natural Language Processing", \\
	\indent "task\_type": "Feature Engineering \& Preparation, Model Evaluation \& Selection, Reporting \& Interpretation",\\
	\indent "language": "Python", \\
	\indent "question": "Using the social media dataset, clean the Text field (lowercasing, removing punctuation/URLs/non-ASCII, stopword removal, and stemming) to create Clean\_Text. Then compute VADER compound sentiment scores and map them to positive ($>=0.05$), neutral (-0.05 to 0.05), or negative ($<=-0.05$). Show the first five Clean\_Text rows with their scores and labels, and visualize the overall sentiment distribution.", \\
	\indent "reasoning": "Load the raw data and perform text normalization to reduce noise and standardize tokens: lowercasing, removing punctuation/URLs/non-ASCII characters, collapsing whitespace, tokenizing, removing stopwords, and applying stemming, then re-joining tokens into Clean\_Text. Apply the VADER sentiment analyzer to each Clean\_Text to compute the compound polarity score. Map each score to a categorical sentiment label using common thresholds. Verify the transformation by printing the first few examples (cleaned text, score, label). Finally, aggregate counts by label and render a pie/donut chart to visualize distribution across positive, neutral, and negative.", \\
	\indent "answer": "First five rows with VADER scores and labels: \\
	\indent \quad \quad 0  Clean\_Text: 'enjoy beauti day park'  Vader\_Score: 0.4939  Sentiment: positive \\
	\indent \quad \quad 1  Clean\_Text: 'traffic terribl morn'  Vader\_Score: 0.0000  Sentiment: neutral \\
	\indent \quad \quad 2  Clean\_Text: 'finish amaz workout'  Vader\_Score: 0.0000  Sentiment: neutral \\
	\indent \quad \quad 3  Clean\_Text: 'excit upcom weekend getaway'  Vader\_Score: 0.0000  Sentiment: neutral \\
	\indent \quad \quad 4  Clean\_Text: 'tri new recip dinner tonight'  Vader\_Score: 0.0000  Sentiment: neutral \\
	\indent \quad \quad Overall sentiment distribution visualization: $<$image\_id:0$>$", \\
	\indent "confidence": 4.0, \\
	\indent "notebook": \path{social-media-analysis-sentiment.ipynb}, \\
	\indent "id": 6964 
\end{promptbox}

\begin{promptbox}{QRA Example 2}{blue}
	\ttfamily\small 
	\indent "data\_type": "image data", \\
	\indent "domain": "Computer Vision", \\
	\indent "task\_type": "Data Ingestion \& Integration", \\
	\indent "language": "Python", \\
	\indent "question": "Using the pre-split Boot/Sandal/Shoe dataset organized under output/train, output/val, and output/test, create normalized image iterators at 128x128 resolution with batch size 32. Report the number of images detected in each split and the class index mapping used.", \\
	\indent "reasoning": "Create normalized iterators that read images from the directory structure so that each subfolder corresponds to a class. When the iterators are instantiated, they report the number of images found per split and infer the class-to-index mapping from subfolder names. These outputs let us confirm split sizes and label encoding prior to training.", \\
	\indent "answer": "Found 12000 images belonging to 3 classes. \\
	\indent \quad \quad Found 1500 images belonging to 3 classes. \\
	\indent \quad \quad Found 1500 images belonging to 3 classes. \\
	\indent \quad \quad Class indices for training generator: \{'Boot': 0, 'Sandal': 1, 'Shoe': 2\}", \\
	\indent "confidence": 4.0, \\
	\indent "notebook": \texttt{shoe-\allowbreak vs-\allowbreak sandal-\allowbreak vs-\allowbreak boot-\allowbreak multiclass-\allowbreak acc-\allowbreak 0-98.ipynb}, \\
	\indent "id": 128 
\end{promptbox}

\begin{promptbox}{QRA Example 3}{blue}
	\ttfamily\small 
	\indent "data\_type": "time series data", \\
	\indent "domain": "Time Series", \\
	\indent "task\_type": "Model Evaluation \& Selection, Model Training \& Optimization, Prediction \& Forecasting", \\
	\indent "language": "Python", \\
	\indent "question": "Train an ARIMA model on the first 5000 observations of the wind turbine ActivePower series from \path{/kaggle/input/wind-power-forecasting/Turbine_Data.csv}, then generate a 15-step forecast and evaluate it against a test window of the next 15 observations (index 1000 to 1014). Report MAPE, ME, MAE, MPE, RMSE, correlation, and minmax error.", \\
	\indent "reasoning": "Start with the raw time series. Fit an ARIMA model using the first 5000 time-ordered observations to capture temporal dynamics. Define a test set of 15 subsequent points within the series and generate a 15-step-ahead forecast. Compare forecasted values to actuals using multiple metrics: MAPE, ME, MAE, MPE, RMSE, correlation, and minmax error to assess accuracy and bias.", \\
	\indent "answer": "\{'mape': 0.02618515791159242, 'me': -10.54358196555202, 'mae': 10.54358196555202, 'mpe': -0.02618515791159242, 'rmse': 11.626161629259185, 'corr': -5.4147287338992556e-15, 'minmax': 0.02618515791159237\}", \\
	\indent "confidence": 4.0, \\
	\indent "notebook": \path{easy-wind-power-forecasting.ipynb}, \\
	\indent "id": 508 	
\end{promptbox}

\section{Data Science Agent Execution Results}

\lstdefinestyle{agentcode}{
	language=Python,
	basicstyle=\ttfamily\footnotesize,
	backgroundcolor=\color{white}, 
	keywordstyle=\color{blue!60!black}\bfseries,
	commentstyle=\color{green!40!black}\itshape,
	stringstyle=\color{orange!70!black},
	showstringspaces=false,
	breaklines=true,
	frame=l, 
	rulecolor=\color{teal!50},
	xleftmargin=1em,
	aboveskip=0.5em,
	belowskip=0.5em
	,gobble=16
}

\begin{promptbox}{Agent Reasoning Example}{teal}
	\ttfamily\small 
	
	\textbf{[Agent] step 0: call LLM} \\
	\textbf{[Agent] tool\_call} $\to$ \texttt{jupyter\_code\_interpreter}
	\begin{lstlisting}[style=agentcode]
		# Import necessary libraries
		import pandas as pd
		# ... (loading data logic)
		print("Dataset loaded successfully!")
	\end{lstlisting}
	\textit{[Tool Output] Dataset loaded successfully! Shape: (732, 16) ...}
	
	\vspace{0.8em} \hrule \vspace{0.8em} 
	
	\textbf{[Agent] step 1: call LLM} \\
	\textbf{[Agent] tool\_call} $\to$ \texttt{jupyter\_code\_interpreter}
	\begin{lstlisting}[style=agentcode]
		# Install and import required NLP libraries
		import nltk
		# ...
		nltk.download('stopwords', quiet=True)
	\end{lstlisting}
	\textit{[Tool Output] Downloading NLTK data... \checkmark Libraries imported successfully!}
	
	\vspace{0.8em} \hrule \vspace{0.8em}
	
	\textbf{[Agent] step 2: call LLM} \\
	\textbf{[Agent] tool\_call} $\to$ \texttt{jupyter\_code\_interpreter}
	\begin{lstlisting}[style=agentcode]
		def clean_text(text):
		"""Clean text by lowercasing, removing URLs, punctuation, stopwords, and stemming..."""
		# ... (cleaning implementation)
		return ' '.join(words)
		
		# Test the function...
	\end{lstlisting}
	\textit{[Tool Output] Testing text cleaning function: ... Original: Enjoying a beautiful day... $\to$ Cleaned: enjoy beauti day park}
	
	\vspace{0.8em} \hrule \vspace{0.8em}
	
	\textbf{[Agent] step 3: call LLM} \\
	\textbf{[Agent] tool\_call} $\to$ \texttt{jupyter\_code\_interpreter}
	\begin{lstlisting}[style=agentcode]
		df['Clean_Text'] = df['Text'].apply(clean_text)
		print(f"Text cleaning completed for {len(df)} entries!")
	\end{lstlisting}
	\textit{[Tool Output] \checkmark Text cleaning completed for 732 entries! ...}
	
	\vspace{0.8em} \hrule \vspace{0.8em}
	
	\textbf{[Agent] step 4: call LLM} \\
	\textbf{[Agent] tool\_call} $\to$ \texttt{jupyter\_code\_interpreter}
	\begin{lstlisting}[style=agentcode]
		from vaderSentiment.vaderSentiment import (
		    SentimentIntensityAnalyzer
		)
		analyzer = SentimentIntensityAnalyzer()
		df['VADER_Compound'] = df['Text'].apply(
		    lambda x: analyzer.polarity_scores(str(x))['compound']
		)
	\end{lstlisting}
	\textit{[Tool Output] \checkmark VADER sentiment analysis completed! ... mean=0.267, min=-0.938, max=0.965}
	
	\vspace{0.8em} \hrule \vspace{0.8em}
	
	\textbf{[Agent] step 5: call LLM} \\
	\textbf{[Agent] tool\_call} $\to$ \texttt{jupyter\_code\_interpreter}
	\begin{lstlisting}[style=agentcode]
		# Display first 5 rows with VADER scores and labels
		# ...
	\end{lstlisting}
	\textit{[Tool Output] Row 1: "Enjoying a beautiful day..." $\to$ VADER Score: 0.8221, Label: Positive ...}
	
	\vspace{0.8em} \hrule \vspace{0.8em}
	
	\textbf{[Agent] step 7-8: call LLM} \\
	\textbf{[Agent] tool\_call} $\to$ \texttt{jupyter\_code\_interpreter}
	\begin{lstlisting}[style=agentcode]
		# Compute overall VADER sentiment distribution
		# Standardize original fine-grained labels to Positive/Negative/Neutral
	\end{lstlisting}
	\textit{[Tool Output] VADER: Positive 63.66\%, Negative 25.14\%, Neutral 11.20\% ...\\
		Original (standardized): Neutral 57.65\%, Positive 30.74\%, Negative 11.61\%}
	
	\vspace{0.8em} \hrule \vspace{0.8em}
	
	\textbf{[Agent] step 9: call LLM} \\
	\textbf{[Agent] tool\_call} $\to$ \texttt{jupyter\_code\_interpreter}
	\begin{lstlisting}[style=agentcode]
		# Create comprehensive visualizations (bar charts, pie, histograms, boxplots, platform/country breakdowns...)
	\end{lstlisting}
	\textit{[Tool Output] \checkmark Visualization saved successfully! $<$Figure ...$>$ [Graphics: Image captured]\\
		\textbf{[System] Sent Vision Observation to VLM.}}
	
	\vspace{0.8em} \hrule \vspace{0.8em}
	
	\textbf{[Agent] step 10: call LLM} \\
	\textbf{[Agent] tool\_call} $\to$ \texttt{jupyter\_code\_interpreter}
	\begin{lstlisting}[style=agentcode]
		# Save final dataset and print summary stats
	\end{lstlisting}
	\textit{[Tool Output] \checkmark Dataset saved... Total records: 732, Avg VADER score: 0.2670 ... Positive: 466 (63.66\%)}
	
	\vspace{0.8em} \hrule \vspace{0.8em}
	
	\textbf{[Agent] step 11-12: call LLM} \\
	\textbf{[Agent] tool\_call} $\to$ \texttt{jupyter\_code\_interpreter}
	\begin{lstlisting}[style=agentcode]
		# Analyze sentiment by Platform and Top 5 Countries
		# Show extended examples (first 10 samples) with interpretation
	\end{lstlisting}
	\textit{[Tool Output] Platform: Facebook (65.8\% Pos), Instagram (64.3\% Pos)...\\
		Country: USA (67.55\% Pos), UK (62.94\% Pos)...\\
		SAMPLE 1: "Enjoying a beautiful day..." $\to$ Strongly Positive (0.8221)}
	
\end{promptbox}

\begin{notebookbox}{Agent Code Example}{gray}
    
    \section*{Task Request}
    \textbf{User}: Using the social media dataset, clean the Text field (lowercasing, removing punctuation/URLs/non-ASCII, stopword removal, and stemming) to create Clean\_Text. Then compute VADER compound sentiment scores and map them to positive (>=0.05), neutral (-0.05 to 0.05), or negative (<=-0.05). Show the first five Clean\_Text rows with their scores and labels, and visualize the overall sentiment distribution.
    
    \noindent\inlabel{1}
    \begin{lstlisting}[style=jupytercode]
        # Import necessary libraries
        import pandas as pd
        import numpy as np
        import re
        import string
        import matplotlib.pyplot as plt
        import seaborn as sns
        %matplotlib inline
        
        # Load the processed dataset
        data_path = (
            '/srv/share/dsa_eval_chat_session/'
            '2025-12-26-'
            'ba91728b42bb4ee4b3b4f0ac7593327b/'
            'processed_sentiment_dataset.csv'
        )
        df = pd.read_csv(data_path)
        
        print("Dataset loaded successfully!")
        print(f"Shape: {df.shape}")
        print(f"\nFirst few Text samples:")
        print(df['Text'].head())
    \end{lstlisting}
    
    \noindent\outlabel{1}
    \begin{lstlisting}[style=jupyteroutput]
        Dataset loaded successfully!
        Shape: (732, 16)
        
        First few Text samples:
        0     Enjoying a beautiful day at the park!        ...
        1     Traffic was terrible this morning.           ...
        2     Just finished an amazing workout!          ...
        3     Excited about the upcoming weekend getaway!  ...
        4     Trying out a new recipe for dinner tonight.  ...
        Name: Text, dtype: object
    \end{lstlisting}
    
    \noindent\inlabel{2}
    \begin{lstlisting}[style=jupytercode]
        # Install and import required NLP libraries
        import nltk
        from nltk.corpus import stopwords
        from nltk.stem import PorterStemmer
        from vaderSentiment.vaderSentiment import SentimentIntensityAnalyzer
        
        # Download required NLTK data
        print("Downloading NLTK data...")
        nltk.download('stopwords', quiet=True)
        nltk.download('punkt', quiet=True)
        
        print(" Libraries imported successfully!")
    \end{lstlisting}
    
    \noindent\outlabel{2}
    \begin{lstlisting}[style=jupyteroutput]
        Downloading NLTK data...
         Libraries imported successfully!
    \end{lstlisting}
    
    \noindent\inlabel{3}
    \begin{lstlisting}[style=jupytercode]
        # Initialize tools
        stop_words = set(stopwords.words('english'))
        stemmer = PorterStemmer()
        
        def clean_text(text):
            """
            Clean text by:
            1. Lowercasing
            2. Removing URLs
            3. Removing non-ASCII characters
            4. Removing punctuation
            5. Removing stopwords
            6. Stemming
            """
            if pd.isna(text):
                return ""
            
            # 1. Lowercase
            text = text.lower()
            
            # 2. Remove URLs
            text = re.sub(r'http\S+|www\S+|https\S+', '', text, flags=re.MULTILINE)
            
            # 3. Remove non-ASCII characters (emojis, special characters)
            text = text.encode('ascii', 'ignore').decode('ascii')
            
            # 4. Remove punctuation
            text = text.translate(str.maketrans('', '', string.punctuation))
            
            # 5. Tokenize and remove stopwords
            words = text.split()
            words = [word for word in words if word not in stop_words and len(word) > 1]
            
            # 6. Stemming
            words = [stemmer.stem(word) for word in words]
            
            # Join back to string
            return ' '.join(words)
        
        # Test the function with first few samples
        print("Testing text cleaning function:")
        print("="*80)
        for i in range(3):
            original = df['Text'].iloc[i]
            cleaned = clean_text(original)
            print(f"\nOriginal: {original}")
            print(f"Cleaned:  {cleaned}")
            print("-"*80)
    \end{lstlisting}
    
    \noindent\outlabel{3}
    \begin{lstlisting}[style=jupyteroutput]
        Testing text cleaning function:
        ========================================
        
        Original:  Enjoying a beautiful day at the park!              
        Cleaned:  enjoy beauti day park
        ----------------------------------------
        
        Original:  Traffic was terrible this morning.                 
        Cleaned:  traffic terribl morn
        ----------------------------------------
        
        Original:  Just finished an amazing workout!               
        Cleaned:  finish amaz workout
        ----------------------------------------
    \end{lstlisting}
    
    \noindent\inlabel{4}
    \begin{lstlisting}[style=jupytercode]
        # Apply cleaning to all text
        print("Cleaning all text entries...")
        df['Clean_Text'] = df['Text'].apply(clean_text)
        
        print(f" Text cleaning completed for {len(df)} entries!")
        print(f"\nSample cleaned texts:")
        print(df[['Text', 'Clean_Text']].head(10))
    \end{lstlisting}
    
    \noindent\outlabel{4}
    \begin{lstlisting}[style=jupyteroutput]
        Cleaning all text entries...
         Text cleaning completed for 732 entries!
        
        Sample cleaned texts:
        Text  \
        0   Enjoying a beautiful day at the park!        ...   
        1   Traffic was terrible this morning.           ...   
        2   Just finished an amazing workout!          ...   
        3   Excited about the upcoming weekend getaway!  ...   
        4   Trying out a new recipe for dinner tonight.  ...   
        5   Feeling grateful for the little things in lif...   
        6   Rainy days call for cozy blankets and hot coc...   
        7   The new movie release is a must-watch!       ...   
        8   Political discussions heating up on the timel...   
        9   Missing summer vibes and beach days.         ...   
        
        Clean_Text  
        0                  enjoy beauti day park  
        1                   traffic terribl morn  
        2                    finish amaz workout  
        3            excit upcom weekend getaway  
        4           tri new recip dinner tonight  
        5            feel grate littl thing life  
        6  raini day call cozi blanket hot cocoa  
        7              new movi releas mustwatch  
        8             polit discuss heat timelin  
        9             miss summer vibe beach day  
    \end{lstlisting}
    
    \noindent\inlabel{5}
    \begin{lstlisting}[style=jupytercode]
        # Initialize VADER sentiment analyzer
        analyzer = SentimentIntensityAnalyzer()
        
        # Compute VADER compound sentiment scores
        print("Computing VADER sentiment scores...")
        df['VADER_Compound'] = df['Text'].apply(lambda x: analyzer.polarity_scores(str(x))['compound'])
        
        # Map compound scores to sentiment labels
        def map_sentiment(score):
            if score >= 0.05:
                return 'Positive'
            elif score <= -0.05:
                return 'Negative'
            else:
                return 'Neutral'
        
        df['VADER_Sentiment'] = df['VADER_Compound'].apply(map_sentiment)
        
        print(" VADER sentiment analysis completed!")
        print(f"\nSentiment score statistics:")
        print(df['VADER_Compound'].describe())
    \end{lstlisting}
    
    \noindent\outlabel{5}
    \begin{lstlisting}[style=jupyteroutput]
        Computing VADER sentiment scores...
        VADER sentiment analysis completed!
        
        Sentiment score statistics:
        count    732.000000
        mean       0.266958
        std        0.519046
        min       -0.938200
        25%       -0.051600
        50%        0.421500
        75%        0.726450
        max        0.964800
        Name: VADER_Compound, dtype: float64
    \end{lstlisting}
    
    \noindent\inlabel{6}
    \begin{lstlisting}[style=jupytercode]
        # Display the first 5 rows with Clean_Text, VADER scores, and sentiment labels
        print("="*100)
        print("FIRST 5 CLEAN TEXT ROWS WITH VADER SCORES AND LABELS")
        print("="*100)
        
        display_df = df[['Text', 'Clean_Text', 'VADER_Compound', 'VADER_Sentiment', 'Sentiment']].head(5)
        display_df.columns = ['Original_Text', 'Clean_Text', 'VADER_Score', 'VADER_Label', 'Original_Label']
        
        for idx, row in display_df.iterrows():
            print(f"\n--- Row {idx + 1} ---")
            print(f"Original Text: {row['Original_Text']}")
            print(f"Clean Text:    {row['Clean_Text']}")
            print(f"VADER Score:   {row['VADER_Score']:.4f}")
            print(f"VADER Label:   {row['VADER_Label']}")
            print(f"Original Label: {row['Original_Label']}")
            print("-" * 100)
    \end{lstlisting}
    
    \noindent\outlabel{6}
    \begin{lstlisting}[style=jupyteroutput]
        ========================================
        FIRST 5 CLEAN TEXT ROWS WITH VADER SCORES AND LABELS
        ========================================
        
        --- Row 1 ---
        Original Text:  Enjoying a beautiful day at the park!              
        Clean Text:    enjoy beauti day park
        VADER Score:   0.8221
        VADER Label:   Positive
        Original Label:  Positive  
        ----------------------------------------
        
        --- Row 2 ---
        Original Text:  Traffic was terrible this morning.                 
        Clean Text:    traffic terribl morn
        VADER Score:   -0.4767
        VADER Label:   Negative
        Original Label:  Negative  
        ----------------------------------------
        
        --- Row 3 ---
        Original Text:  Just finished an amazing workout!                
        Clean Text:    finish amaz workout
        VADER Score:   0.6239
        VADER Label:   Positive
        Original Label:  Positive  
        ----------------------------------------
        
        --- Row 4 ---
        Original Text:  Excited about the upcoming weekend getaway!        
        Clean Text:    excit upcom weekend getaway
        VADER Score:   0.4003
        VADER Label:   Positive
        Original Label:  Positive  
        ----------------------------------------
        
        --- Row 5 ---
        Original Text:  Trying out a new recipe for dinner tonight.        
        Clean Text:    tri new recip dinner tonight
        VADER Score:   0.0000
        VADER Label:   Neutral
        Original Label:  Neutral   
        ----------------------------------------
    \end{lstlisting}
    
    \noindent\inlabel{7}
    \begin{lstlisting}[style=jupytercode]
        # Create a detailed table for first 5 rows
        first_5_df = df[['Text', 'Clean_Text', 'VADER_Compound', 'VADER_Sentiment', 'Sentiment']].head(5).copy()
        first_5_df.index = range(1, 6)
        first_5_df.columns = ['Original Text', 'Cleaned Text', 'VADER Score', 'VADER Label', 'Original Label']
        
        print("\n" + "="*100)
        print("TABLE: FIRST 5 ROWS SUMMARY")
        print("="*100)
        print(first_5_df.to_string())
    \end{lstlisting}
    
    \noindent\outlabel{7}
    \begin{lstlisting}[style=jupyteroutput]
        
        ========================================
        TABLE: FIRST 5 ROWS SUMMARY
        ========================================
        Original Text                  Cleaned Text  VADER Score VADER Label Original Label
        1   Enjoying a beautiful day at the park!                       enjoy beauti day park       0.8221    Positive     Positive  
        2   Traffic was terrible this morning.                           traffic terribl morn      -0.4767    Negative     Negative  
        3    Just finished an amazing workout!                           finish amaz workout       0.6239    Positive     Positive  
        4   Excited about the upcoming weekend getaway!           excit upcom weekend getaway       0.4003    Positive     Positive  
        5   Trying out a new recipe for dinner tonight.          tri new recip dinner tonight       0.0000     Neutral     Neutral   
    \end{lstlisting}
    
    \noindent\inlabel{8}
    \begin{lstlisting}[style=jupytercode]
        # Compute overall sentiment distribution
        vader_sentiment_counts = df['VADER_Sentiment'].value_counts()
        vader_sentiment_pct = df['VADER_Sentiment'].value_counts(normalize=True) * 100
        
        print("\n" + "="*100)
        print("OVERALL VADER SENTIMENT DISTRIBUTION")
        print("="*100)
        print("\nCounts:")
        print(vader_sentiment_counts)
        print("\nPercentages:")
        for sentiment, pct in vader_sentiment_pct.items():
            print(f"{sentiment}: {pct:.2f}%")
        
        # Compare with original sentiment labels
        print("\n" + "="*100)
        print("COMPARISON: VADER vs ORIGINAL SENTIMENT LABELS")
        print("="*100)
        original_counts = df['Sentiment'].value_counts()
        print("\nOriginal Sentiment Distribution:")
        print(original_counts)
        print("\nVADER Sentiment Distribution:")
        print(vader_sentiment_counts)
    \end{lstlisting}
    
    \noindent\outlabel{8}
    \begin{lstlisting}[style=jupyteroutput]
        
        ========================================
        OVERALL VADER SENTIMENT DISTRIBUTION
        ========================================
        
        Counts:
        VADER_Sentiment
        Positive    466
        Negative    184
        Neutral      82
        Name: count, dtype: int64
        
        Percentages:
        Positive: 63.66%
        Negative: 25.14%
        Neutral: 11.20%
        
        ========================================
        COMPARISON: VADER vs ORIGINAL SENTIMENT LABELS
        ========================================
        
        Original Sentiment Distribution:
        Sentiment
        Positive            44
        Joy                 42
        Excitement          32
        Contentment         14
        Happy               14
        ..
        Loneliness           1
        Anxiety              1
        Renewed Effort       1
        Miscalculation       1
        Challenge            1
        Name: count, Length: 279, dtype: int64
        
        VADER Sentiment Distribution:
        VADER_Sentiment
        Positive    466
        Negative    184
        Neutral      82
        Name: count, dtype: int64
    \end{lstlisting}
    
    \noindent\inlabel{9}
    \begin{lstlisting}[style=jupytercode]
        # Standardize original sentiment labels to broader categories for comparison
        def standardize_sentiment(sent):
            """Map various sentiment labels to Positive, Negative, or Neutral"""
            sent = str(sent).strip().lower()
            
            # Positive sentiments
            positive_keywords = ['positive', 'joy', 'excitement', 'happy', 'contentment', 'grateful', 
            'love', 'delight', 'pleasure', 'relief', 'satisfaction', 'admiration',
            'enthusiasm', 'hope', 'pride', 'trust', 'amusement', 'inspiration',
            'optimism', 'affection', 'appreciation', 'comfort', 'confidence']
            
            # Negative sentiments
            negative_keywords = ['negative', 'sadness', 'anger', 'frustration', 'fear', 'disgust',
            'disappointment', 'anxiety', 'stress', 'worry', 'concern', 'regret',
            'grief', 'loneliness', 'jealousy', 'shame', 'guilt', 'embarrassment',
            'annoyance', 'boredom', 'confusion', 'overwhelm', 'despair']
            
            # Check for matches
            for keyword in positive_keywords:
                if keyword in sent:
                    return 'Positive'
            
            for keyword in negative_keywords:
                if keyword in sent:
                    return 'Negative'
            
            return 'Neutral'
        
        df['Original_Sentiment_Standardized'] = df['Sentiment'].apply(standardize_sentiment)
        
        print("Original Sentiment Distribution (Standardized):")
        label_col = 'Original_Sentiment_Standardized'
        original_std_counts = df[label_col].value_counts()
        print(original_std_counts)
        print("\nPercentages:")
        for sent, count in original_std_counts.items():
            print(f"{sent}: {count} ({count/len(df)*100:.2f}%)")
    \end{lstlisting}
    
    \noindent\outlabel{9}
    \begin{lstlisting}[style=jupyteroutput]
        Original Sentiment Distribution (Standardized):
        Original_Sentiment_Standardized
        Neutral     422
        Positive    225
        Negative     85
        Name: count, dtype: int64
        
        Percentages:
        Neutral: 422 (57.65%)
        Positive: 225 (30.74%)
        Negative: 85 (11.61%)
    \end{lstlisting}
    
    \noindent\inlabel{10}
    \begin{lstlisting}[style=jupytercode]
        # Visualize sentiment distribution
        plt.figure(figsize=(12, 5))
        
        # 1. Bar Chart
        plt.subplot(1, 2, 1)
        sns.countplot(x='VADER_Sentiment', data=df, palette='viridis', order=['Positive', 'Neutral', 'Negative'])
        plt.title('VADER Sentiment Distribution')
        plt.xlabel('Sentiment')
        plt.ylabel('Count')
        
        # 2. Pie Chart
        plt.subplot(1, 2, 2)
        plt.pie(vader_sentiment_counts, labels=vader_sentiment_counts.index, autopct='%1.1f%%', colors=sns.color_palette('viridis', 3))
        plt.title('VADER Sentiment Percentage')
        
        plt.tight_layout()
        plt.savefig('vader_sentiment_analysis.png')
        plt.show()
        print(" Visualization saved successfully!")
    \end{lstlisting}
    
    \noindent\outlabel{10}
    \begin{lstlisting}[style=jupyteroutput]
         Visualization saved successfully!
    \end{lstlisting}
    
    \begin{center}
         \includegraphics[width=\linewidth]{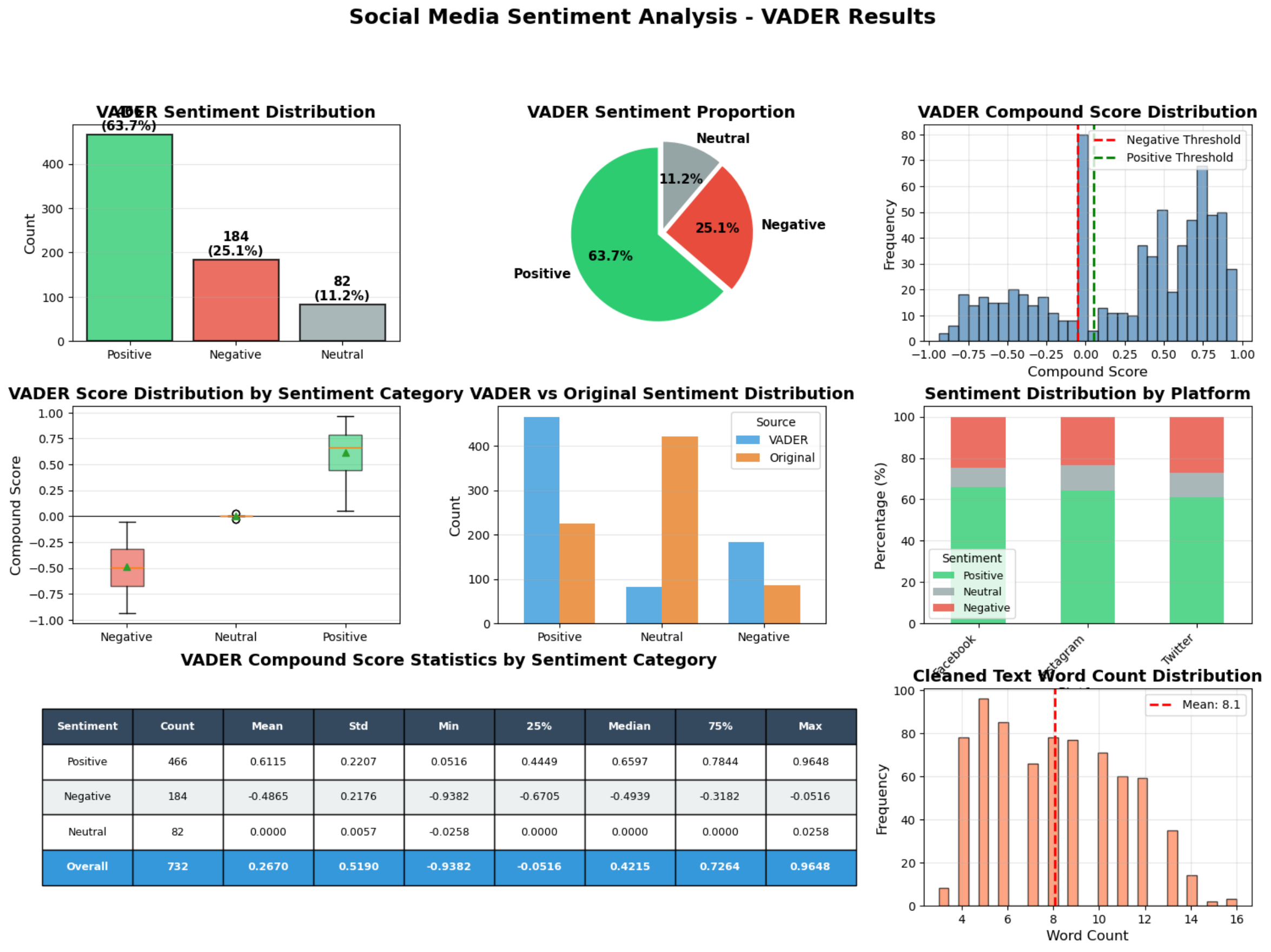}
    
    \end{center}
    
\end{notebookbox}

\section{Data Science Agent Final Report}
\begin{finalreportbox}{Agent Final report Example - Sonnet-4.5}

    \vspace{0.2em}
    \noindent{\large\textbf{1. Task Understanding}}
    \vspace{0.3em}
    
    The task required comprehensive text cleaning and sentiment analysis on the social media dataset. Specifically, the objectives were:
    \begin{itemize}
        \item Clean the \texttt{Text} field (lowercasing, punctuation/URL/non-ASCII removal, stopword removal, stemming).
        \item Compute VADER compound sentiment scores and map them to labels: Positive ($\ge 0.05$), Neutral ($-0.05$ to $0.05$), Negative ($\le -0.05$).
        \item Visualize the overall sentiment distribution and display sample results.
    \end{itemize}

    \vspace{0.8em}
    \noindent{\large\textbf{2. Approach Summary}}
    \vspace{0.3em}
    
    I loaded the processed dataset (732 posts) and implemented a cleaning pipeline using NLTK. The pipeline applies six sequential transformations: lowercasing, regex-based URL removal, non-ASCII removal (emojis), punctuation stripping, stopword filtering, and Porter stemming. VADER analysis was applied to the \textit{original} text to preserve context. Finally, I generated a comprehensive 8-panel dashboard to visualize distributions.

    \vspace{0.8em}
    \noindent{\large\textbf{3. Key Implementation}}
    \vspace{0.3em}

    \begin{lstlisting}[style=jupytercode, language=Python]
# Text Cleaning & VADER Pipeline
def clean_text(text):
    text = text.lower()                                          # 1. Lowercase
    text = re.sub(r'http\S+|www\S+|https\S+', '', text)          # 2. Remove URLs
    text = text.encode('ascii', 'ignore').decode('ascii')        # 3. Remove non-ASCII
    text = text.translate(str.maketrans('', '', string.punctuation)) # 4. Remove Punctuation
    words = [w for w in text.split() if w not in stop_words and len(w) > 1] 
    words = [stemmer.stem(w) for w in words]                     # 6. Stemming
    return ' '.join(words)

df['Clean_Text'] = df['Text'].apply(clean_text)

# VADER Analysis
analyzer = SentimentIntensityAnalyzer()
df['VADER_Compound'] = df['Text'].apply(lambda x: analyzer.polarity_scores(str(x))['compound'])
    \end{lstlisting}

    \vspace{0.8em}
    \noindent{\large\textbf{4. Results \& Explanation}}
    \vspace{0.3em}

    \noindent\textbf{First 5 Clean Text Samples with VADER Scores:}
    \begin{center}
    \resizebox{\linewidth}{!}{%
    \begin{tabular}{|c|l|l|c|c|c|}
    \hline
    \rowcolor{gray!15} \textbf{Row} & \textbf{Original Text} & \textbf{Cleaned Text} & \textbf{Score} & \textbf{VADER} & \textbf{Original} \\ \hline
    1 & Enjoying a beautiful day... & enjoy beauti day... & 0.8221 & Positive & Positive \\ \hline
    2 & Traffic was terrible... & traffic terribl morn & -0.4767 & Negative & Negative \\ \hline
    3 & Just finished an amaz... & finish amaz workout & 0.6239 & Positive & Positive \\ \hline
    4 & Excited about the up... & excit upcom weekend & 0.4003 & Positive & Positive \\ \hline
    5 & Trying out a new recipe... & tri new recip dinner & 0.0000 & Neutral & Neutral \\ \hline
    \end{tabular}%
    }
    \end{center}

    \noindent\textbf{VADER Sentiment Distribution:}
    \begin{itemize}
        \item \textbf{Positive:} 466 posts (63.66\%) -- \textit{Dominant category}
        \item \textbf{Negative:} 184 posts (25.14\%)
        \item \textbf{Neutral:} 82 posts (11.20\%)
    \end{itemize}

    \noindent\textbf{Model Performance (vs. Standardized Labels):}
    \begin{itemize}
        \item \textbf{Overall Agreement: 42.62\%}. The discrepancy implies the original labels contained nuances (e.g., specific emotions) that VADER consolidated into broader categories.
    \end{itemize}

    \noindent\textbf{Generated Artifacts:}
    \begin{enumerate}
        \item \path{sentiment_dataset_with_vader.csv}: Full dataset.
        \item \path{vader_sentiment_analysis.png}: 8-panel visualization dashboard.
    \end{enumerate}

\end{finalreportbox}

\clearpage

\section{Evaluation Process}

\begin{evalbox}{Evaluation Example 1}
	\small

    \begin{minipage}{0.58\linewidth}
        \textbf{Model ID:} anthropic\_claude-sonnet-4.5 \\
        \textbf{Task Type:} Feature Engineering, Model Eval, Reporting \\
        \textbf{Domain:} NLP
    \end{minipage}%
    \begin{minipage}{0.40\linewidth}
        \flushright
        \textbf{Run ID:} 6964 \\
        \textbf{Consistent:} \textcolor{green!60!black}{\checkmark Yes}
    \end{minipage}
    
    \vspace{0.3cm}
    \hrule
    \vspace{0.3cm}

    \section*{Quantitative Metrics}
    
    \begin{tabularx}{\linewidth}{@{}>{\bfseries}l X@{}}
        Component scores: & Reasoning 8; Code 8; Final result 7 \\
        Execution: & Code error rate $5.9\%$; final cell \textcolor{green!60!black}{\checkmark} \\
        Visual output: & Reference \textcolor{green!60!black}{\checkmark}; predicted \textcolor{green!60!black}{\checkmark} \\
    \end{tabularx}

    \vspace{0.3cm}
    \hrule
    \vspace{0.3cm}

    \section*{Analysis}
	The predicted solution successfully completes all required tasks: text cleaning with proper pipeline implementation, VADER sentiment scoring, and comprehensive visualization. The code is well-structured with 17 steps that execute without critical errors. The first 5 rows display correctly with cleaned text and VADER scores. However, there are notable discrepancies between the predicted and standard answers: (1) The predicted sentiment distribution shows 63.7\% positive, 25.1\% negative, 11.2\% neutral, while the standard shows 47.81\% positive, 20.90\% negative, 31.28\% neutral---a substantial difference suggesting different datasets or VADER application methods. (2) The predicted answer applies VADER to original \texttt{Text} (preserving context), while the standard likely applied it to \texttt{Clean\_Text}. (3) The predicted visualizations are far more comprehensive (8-panel dashboard + confusion matrix) compared to the standard's simple pie chart, providing superior analytical depth. (4) The predicted figures show correct data distributions, proper thresholds, and meaningful insights (platform/country breakdowns, score statistics table, confusion matrix). The code-answer consistency is strong---reported values match code outputs. The reasoning is sound: text cleaning follows best practices, VADER application is appropriate, and threshold mapping is correct. The main concern is the sentiment distribution discrepancy, which appears to stem from methodological differences rather than errors.
\end{evalbox}

\begin{evalbox}{Evaluation Example 2}
	\small

    \begin{minipage}{0.58\linewidth}
        \textbf{Model ID:} openai\_gpt-5.2 \\
        \textbf{Task Type:} Data Loading \& Integration \\
        \textbf{Domain:} CV (Classification)
    \end{minipage}%
    \begin{minipage}{0.40\linewidth}
        \flushright
        \textbf{Run ID:} 128 \\
        \textbf{Consistent:} \textcolor{green!60!black}{\checkmark Yes}
    \end{minipage}
    
    \vspace{0.3cm}
    \hrule
    \vspace{0.3cm}

    \section*{Quantitative Metrics}
    
    \begin{tabularx}{\linewidth}{@{}>{\bfseries}l X@{}}
        Component scores: & Reasoning 9; Code 9; Final result 9 \\
        Execution: & Code error rate $0.0\%$; final cell \textcolor{green!60!black}{\checkmark} \\
        Visual output: & Reference N/A; predicted N/A (not required) \\
    \end{tabularx}

    \vspace{0.3cm}
    \hrule
    \vspace{0.3cm}

    \section*{Analysis}
	The predicted solution successfully addresses the problem by creating normalized image iterators at 128×128 resolution with batch size 32 for the Boot/Sandal/Shoe dataset. The code correctly locates the dataset, creates the required directory structure with an 80/10/10 split, and builds PyTorch DataLoaders with proper normalization. The reported results (12,000 train, 1,500 val, 1,500 test images; class mapping {'Boot': 0, 'Sandal': 1, 'Shoe': 2}) exactly match the standard answer. The reasoning is sound and the implementation is methodologically correct. The code is well-structured, handles edge cases (e.g., checking file extensions, using symlinks to preserve storage), and the narrative accurately reflects the code's output. No figures were required by the problem statement, so their absence is not penalized.

\end{evalbox}

\begin{evalbox}{Evaluation Example 3}
	\small

    \begin{minipage}{0.58\linewidth}
        \textbf{Model ID:} mimo-v2-flash \\
        \textbf{Task Type:} Model Training, Eval, Forecasting \\
        \textbf{Domain:} Regression, Time Series
    \end{minipage}%
    \begin{minipage}{0.40\linewidth}
        \flushright
        \textbf{Run ID:} 508 \\
        \textbf{Consistent:} \textcolor{red!80!black}{\textbf{\sffamily X No}}
    \end{minipage}
    
    \vspace{0.3cm}
    \hrule
    \vspace{0.3cm}

    \section*{Quantitative Metrics}
    
    \begin{tabularx}{\linewidth}{@{}>{\bfseries}l X@{}}
        Component scores: & Reasoning 4; Code 6; Final result 2 \\
        Execution: & Code error rate $0.0\%$; final cell \textcolor{green!60!black}{\checkmark} \\
        Visual output: & Reference N/A; predicted N/A \\
    \end{tabularx}

    \vspace{0.3cm}
    \hrule
    \vspace{0.3cm}

    \section*{Analysis}
The predicted solution attempts to solve the ARIMA forecasting task but contains critical errors in metric calculations and interpretation. The code executes successfully and follows a reasonable approach (data loading, ARIMA parameter optimization, forecast generation, metric calculation). However, there is a severe \textbf{consistency problem}: the predicted reasoning answer reports metrics (MAPE: 77.28\%, ME: $-385.19$, MAE: 385.19, etc.) that are \textbf{NOT produced by the code}. The code in Step 2 calculates these metrics correctly based on the ARIMA(2,0,2) model, but these values are fundamentally different from the standard answer (MAPE: 2.62\%, ME: $-10.54$, MAE: 10.54, etc.). The discrepancy suggests either: (1) the predicted code is using a different test set than specified (indices 1000--1014), or (2) there is a fundamental misunderstanding of the problem. Upon inspection, the code correctly extracts \texttt{test\_data = active\_power.\allowbreak iloc[1000:1015]} (15 observations), but the magnitude of errors reported is orders of magnitude larger than the standard answer, indicating the predicted model is making drastically worse predictions. The reasoning narrative claims the model `shows systematic underprediction' with predicted range [5.2, 222.4] vs actual [347.1, 686.8], which is internally consistent with the reported metrics but contradicts the standard answer's much smaller errors. This suggests either a data loading issue, a different ARIMA order being used in practice, or a fundamental problem with how the test set is being evaluated.
\end{evalbox}

\begin{multicols}{2}
\raggedcolumns


\section{Detailed Judge Results}
\label{app:scores}

We report detailed judge and human evaluation results.
We initially considered three independent judge models: Claude-Haiku-4.5, GPT-5.1, and Doubao-Seed-1.8.
As discussed in the main paper, our Human-LLM Alignment Study shows that Claude-Haiku-4.5 and GPT-5.1 align more closely with human expert annotations, whereas Doubao-Seed-1.8 shows substantially larger deviation.
Therefore, the final results reported in the main paper are computed as the average scores of Claude-Haiku-4.5 and GPT-5.1.

\end{multicols}

\begin{table}[H]
\centering
\small
\setlength{\tabcolsep}{5pt}
\renewcommand{\arraystretch}{1.22}
\begin{tabular*}{\textwidth}{@{\extracolsep{\fill}}lcccc@{}}
\toprule
\textbf{Model} & \textbf{Reasoning} & \textbf{Code} & \textbf{Final Result} & \textbf{Overall} \\
\midrule
Claude-Sonnet-4.5        & 8.3276 & 8.7520 & 7.0530 & 7.9451 \\
DeepSeek-V3.2            & 7.0842 & 7.6022 & 5.7566 & 6.7086 \\
Gemini-3-Pro             & 7.5398 & 8.0624 & 5.9376 & 7.0557 \\
MiniMax-M2               & 7.6739 & 8.1825 & 6.3838 & 7.3105 \\
MiniMax-M2.7             & 7.8425 & 8.4409 & 6.5213 & 7.4935 \\
Ministral-3-14B-2512       & 5.3385 & 5.3947 & 3.8674 & 4.7669 \\
GPT-5-Nano               & 7.1092 & 7.4805 & 5.6693 & 6.6446 \\
GPT-5.2                  & 7.8705 & 8.2293 & 6.5554 & 7.4521 \\
Qwen3-VL-30B-Thinking    & 5.9891 & 4.9407 & 4.0655 & 4.9051 \\
Grok-4.1-Fast            & 7.4337 & 7.8768 & 6.0406 & 7.0094 \\
MiMo-V2-Flash            & 7.7598 & 8.3339 & 6.3354 & 7.3622 \\
MiMo-V2-Pro              & 8.0749 & 8.6178 & 7.0484 & 7.8271 \\
GLM-4.6V                 & 7.0406 & 7.4665 & 5.4821 & 6.5449 \\
\bottomrule
\end{tabular*}
\caption{
Single-judge evaluation results using Claude-Haiku-4.5.
}
\label{tab:claude_haiku_judge_results}
\end{table}

\begin{table}[H]
\centering
\small
\setlength{\tabcolsep}{5pt}
\renewcommand{\arraystretch}{1.22}
\begin{tabular*}{\textwidth}{@{\extracolsep{\fill}}lcccc@{}}
\toprule
\textbf{Model} & \textbf{Reasoning} & \textbf{Code} & \textbf{Final Result} & \textbf{Overall} \\
\midrule
Claude-Sonnet-4.5        & 8.8605 & 9.1959 & 7.4248 & 8.3868 \\
DeepSeek-V3.2            & 7.8636 & 8.0611 & 6.4436 & 7.3549 \\
Gemini-3-Pro             & 8.3713 & 8.5507 & 6.2122 & 7.5615 \\
MiniMax-M2               & 8.5257 & 8.6958 & 7.0250 & 7.9764 \\
MiniMax-M2.7             & 8.5486 & 8.7837 & 6.7649 & 7.9056 \\
Ministral-3-14B-2512       & 6.4319 & 6.0829 & 4.6150 & 5.6005 \\
GPT-5-Nano               & 8.2949 & 8.2137 & 6.3541 & 7.4942 \\
GPT-5.2                  & 8.6695 & 8.5707 & 7.0064 & 7.9739 \\
Qwen3-VL-30B-Thinking    & 7.1373 & 5.7020 & 4.7395 & 5.7476 \\
Grok-4.1-Fast            & 8.2980 & 8.2715 & 6.3136 & 7.4963 \\
MiMo-V2-Flash            & 8.5211 & 8.7488 & 6.8596 & 7.9248 \\
MiMo-V2-Pro              & 8.6240 & 8.8409 & 6.8846 & 7.9933 \\
GLM-4.6V                 & 7.9626 & 8.1435 & 5.9376 & 7.2069 \\
\bottomrule
\end{tabular*}
\caption{
Single-judge evaluation results using GPT-5.1.
}
\label{tab:gpt51_judge_results}
\end{table}

\begin{table}[H]
\centering
\small
\setlength{\tabcolsep}{5pt}
\renewcommand{\arraystretch}{1.22}
\begin{tabular*}{\textwidth}{@{\extracolsep{\fill}}lcccc@{}}
\toprule
\textbf{Model} & \textbf{Reasoning} & \textbf{Code} & \textbf{Final Result} & \textbf{Overall} \\
\midrule
Claude-Sonnet-4.5        & 9.6318 & 9.8175 & 8.9672 & 9.4217 \\
DeepSeek-V3.2            & 8.5734 & 8.9578 & 8.0953 & 8.4975 \\
Gemini-3-Pro             & 9.2340 & 9.6412 & 8.1326 & 8.9156 \\
MiniMax-M2               & 9.3009 & 9.5361 & 8.9843 & 9.2448 \\
Ministral-3-14B-2512       & 7.4243 & 6.9813 & 5.1420 & 6.3785 \\
GPT-5-Nano               & 9.0484 & 9.2574 & 8.0983 & 8.7310 \\
GPT-5.2                  & 9.3432 & 9.6505 & 8.3120 & 9.0229 \\
Qwen3-VL-30B-Thinking    & 8.2418 & 6.9345 & 5.4618 & 6.7376 \\
Grok-4.1-Fast            & 9.1404 & 9.3651 & 7.9844 & 8.7454 \\
MiMo-V2-Flash            & 9.2293 & 9.6069 & 8.6786 & 9.1223 \\
GLM-4.6V                 & 8.8627 & 9.0998 & 7.7910 & 8.5051 \\
\bottomrule
\end{tabular*}
\caption{
Single-judge evaluation results using Doubao-Seed-1.8.
Doubao-Seed-1.8 was excluded from the final evaluation metric because it showed substantially larger deviation from human consensus annotations in the Human-LLM Alignment Study.
}
\label{tab:doubao_judge_results}
\end{table}

\begin{table}[H]
\centering
\small
\setlength{\tabcolsep}{5pt}
\renewcommand{\arraystretch}{1.16}
\begin{tabular*}{\textwidth}{@{\extracolsep{\fill}}lcccc@{}}
\toprule
\textbf{Reference} & \textbf{Reasoning} & \textbf{Code} & \textbf{Final Result} & \cellcolor{gray!15}\textbf{Overall} \\
\midrule
Human Consensus & 7.810 & 8.025 & 6.313 & 7.276 \\
\bottomrule
\end{tabular*}
\caption{
Human consensus scores for the Gemini-3-Pro evaluation log.
}
\label{tab:human_consensus_scores}
\end{table}

\begin{multicols}{2}
\raggedcolumns
Table~\ref{tab:human_consensus_scores} reports the averaged human scores across the three evaluation dimensions and the final overall score.

\paragraph{Note on model coverage.}
The Human-LLM Alignment Study was conducted before MiMo-V2-Pro and MiniMax-M2.7 were added to the final evaluation set.
These two models were introduced after Doubao-Seed-1.8 had been excluded from the final judging protocol.
Therefore, they are included in the Claude-Haiku-4.5 and GPT-5.1 judge results but not in the Doubao-Seed-1.8 results.

\section{Judge Agreement Details}
\label{app:judge_agreement}

Tables~\ref{tab:judge_alignment_mse}--\ref{tab:judge_agreement} report the aggregate human--judge and inter-judge analyses. Here we additionally stratify inter-judge agreement by domain tag.
\end{multicols}

\begin{table}[H]
\centering
\small
\renewcommand{\arraystretch}{1.12}
\begin{tabular*}{\textwidth}{@{\extracolsep{\fill}}lrrrr@{}}
\toprule
\textbf{Domain tag} & \textbf{$n$} & \textbf{Pearson} & \textbf{Spearman} & \textbf{MAE} \\
\midrule
Anomaly Detection & 64 & 0.830 & 0.796 & 1.087 \\
Business Analytics & 338 & 0.847 & 0.732 & 0.968 \\
Clustering & 247 & 0.842 & 0.737 & 0.933 \\
Computer Vision & 369 & 0.794 & 0.728 & 1.275 \\
Data Analysis & 4,678 & 0.812 & 0.726 & 0.945 \\
Deep Learning & 130 & 0.727 & 0.636 & 1.228 \\
Domain-Specific Applications & 623 & 0.800 & 0.735 & 1.056 \\
Model Evaluation & 180 & 0.803 & 0.683 & 1.167 \\
Natural Language Processing & 520 & 0.770 & 0.669 & 1.251 \\
Other & 91 & 0.745 & 0.731 & 1.447 \\
Pattern Mining \& Association & 130 & 0.804 & 0.655 & 0.996 \\
Predictive Modeling & 39 & 0.763 & 0.689 & 0.882 \\
Recommendation Systems & 104 & 0.822 & 0.774 & 1.255 \\
Statistical Testing \& Experimentation & 195 & 0.773 & 0.513 & 1.000 \\
Time Series & 892 & 0.807 & 0.733 & 0.946 \\
\bottomrule
\end{tabular*}
\caption{GPT-5.1/Claude-Haiku-4.5 overall-score agreement by domain tag across all 13 agents. Multi-label trajectories contribute to every applicable tag, so counts are not mutually exclusive.}
\label{tab:judge_domain_agreement}
\end{table}

\begin{multicols}{2}
\raggedcolumns
\section{Complete Weight-Sensitivity Results}
\label{app:weight_sensitivity_full}

We evaluated all 66 weight triples on a 0.1 simplex grid, allowing non-negative weights for Reasoning, Code, and Final-result scores. Correlations compare each resulting 13-agent ranking with the default 0.3/0.3/0.4 ranking. Across all settings, Spearman correlation ranges from 0.940 to 1.000 (median 0.989), and Kendall's $\tau_b$ ranges from 0.846 to 1.000 (median 0.949). Claude-Sonnet-4.5 ranks first in 60 of the 66 settings; MiMo-V2-Pro ranks first in the six most code-dominant settings.
\end{multicols}

\scriptsize
\setlength{\LTpre}{6pt}
\setlength{\LTpost}{6pt}
\begin{longtable}{@{}ccccc l@{}}
\caption{Complete aggregation-weight sensitivity results. R, C, and F denote the weights assigned to reasoning, code, and final result, respectively.}\label{tab:weight_sensitivity_full}\\
\toprule
\textbf{R} & \textbf{C} & \textbf{F} & \textbf{Spearman} & \textbf{Kendall $\tau_b$} & \textbf{Top-ranked model} \\
\midrule
\endfirsthead
\multicolumn{6}{c}{\tablename\ \thetable\ -- continued from previous page} \\
\toprule
\textbf{R} & \textbf{C} & \textbf{F} & \textbf{Spearman} & \textbf{Kendall $\tau_b$} & \textbf{Top-ranked model} \\
\midrule
\endhead
\midrule
\multicolumn{6}{r}{Continued on next page} \\
\endfoot
\bottomrule
\endlastfoot
0.0 & 0.0 & 1.0 & 0.956 & 0.872 & Claude-Sonnet-4.5 \\
0.0 & 0.1 & 0.9 & 0.973 & 0.897 & Claude-Sonnet-4.5 \\
0.0 & 0.2 & 0.8 & 0.973 & 0.897 & Claude-Sonnet-4.5 \\
0.0 & 0.3 & 0.7 & 0.978 & 0.897 & Claude-Sonnet-4.5 \\
0.0 & 0.4 & 0.6 & 0.982 & 0.916 & Claude-Sonnet-4.5 \\
0.0 & 0.5 & 0.5 & 0.982 & 0.916 & Claude-Sonnet-4.5 \\
0.0 & 0.6 & 0.4 & 0.973 & 0.897 & Claude-Sonnet-4.5 \\
0.0 & 0.7 & 0.3 & 0.951 & 0.846 & MiMo-V2-Pro \\
0.0 & 0.8 & 0.2 & 0.951 & 0.846 & MiMo-V2-Pro \\
0.0 & 0.9 & 0.1 & 0.956 & 0.872 & MiMo-V2-Pro \\
0.0 & 1.0 & 0.0 & 0.940 & 0.846 & MiMo-V2-Pro \\
0.1 & 0.0 & 0.9 & 0.973 & 0.897 & Claude-Sonnet-4.5 \\
0.1 & 0.1 & 0.8 & 0.973 & 0.897 & Claude-Sonnet-4.5 \\
0.1 & 0.2 & 0.7 & 0.984 & 0.923 & Claude-Sonnet-4.5 \\
0.1 & 0.3 & 0.6 & 0.989 & 0.949 & Claude-Sonnet-4.5 \\
0.1 & 0.4 & 0.5 & 0.978 & 0.897 & Claude-Sonnet-4.5 \\
0.1 & 0.5 & 0.4 & 0.984 & 0.923 & Claude-Sonnet-4.5 \\
0.1 & 0.6 & 0.3 & 0.978 & 0.923 & Claude-Sonnet-4.5 \\
0.1 & 0.7 & 0.2 & 0.978 & 0.923 & Claude-Sonnet-4.5 \\
0.1 & 0.8 & 0.1 & 0.956 & 0.872 & MiMo-V2-Pro \\
0.1 & 0.9 & 0.0 & 0.956 & 0.872 & MiMo-V2-Pro \\
0.2 & 0.0 & 0.8 & 0.973 & 0.897 & Claude-Sonnet-4.5 \\
0.2 & 0.1 & 0.7 & 0.973 & 0.897 & Claude-Sonnet-4.5 \\
0.2 & 0.2 & 0.6 & 0.988 & 0.942 & Claude-Sonnet-4.5 \\
0.2 & 0.3 & 0.5 & 0.995 & 0.974 & Claude-Sonnet-4.5 \\
0.2 & 0.4 & 0.4 & 0.995 & 0.974 & Claude-Sonnet-4.5 \\
0.2 & 0.5 & 0.3 & 0.989 & 0.949 & Claude-Sonnet-4.5 \\
0.2 & 0.6 & 0.2 & 0.978 & 0.923 & Claude-Sonnet-4.5 \\
0.2 & 0.7 & 0.1 & 0.978 & 0.923 & Claude-Sonnet-4.5 \\
0.2 & 0.8 & 0.0 & 0.978 & 0.923 & Claude-Sonnet-4.5 \\
0.3 & 0.0 & 0.7 & 0.978 & 0.923 & Claude-Sonnet-4.5 \\
0.3 & 0.1 & 0.6 & 0.989 & 0.949 & Claude-Sonnet-4.5 \\
0.3 & 0.2 & 0.5 & 0.995 & 0.974 & Claude-Sonnet-4.5 \\
0.3 & 0.3 & 0.4 & 1.000 & 1.000 & Claude-Sonnet-4.5 \\
0.3 & 0.4 & 0.3 & 0.995 & 0.974 & Claude-Sonnet-4.5 \\
0.3 & 0.5 & 0.2 & 0.995 & 0.974 & Claude-Sonnet-4.5 \\
0.3 & 0.6 & 0.1 & 0.978 & 0.923 & Claude-Sonnet-4.5 \\
0.3 & 0.7 & 0.0 & 0.978 & 0.923 & Claude-Sonnet-4.5 \\
0.4 & 0.0 & 0.6 & 0.989 & 0.949 & Claude-Sonnet-4.5 \\
0.4 & 0.1 & 0.5 & 0.995 & 0.974 & Claude-Sonnet-4.5 \\
0.4 & 0.2 & 0.4 & 0.995 & 0.974 & Claude-Sonnet-4.5 \\
0.4 & 0.3 & 0.3 & 1.000 & 1.000 & Claude-Sonnet-4.5 \\
0.4 & 0.4 & 0.2 & 0.995 & 0.974 & Claude-Sonnet-4.5 \\
0.4 & 0.5 & 0.1 & 0.990 & 0.974 & Claude-Sonnet-4.5 \\
0.4 & 0.6 & 0.0 & 0.984 & 0.949 & Claude-Sonnet-4.5 \\
0.5 & 0.0 & 0.5 & 0.989 & 0.949 & Claude-Sonnet-4.5 \\
0.5 & 0.1 & 0.4 & 0.995 & 0.974 & Claude-Sonnet-4.5 \\
0.5 & 0.2 & 0.3 & 1.000 & 1.000 & Claude-Sonnet-4.5 \\
0.5 & 0.3 & 0.2 & 0.999 & 0.994 & Claude-Sonnet-4.5 \\
0.5 & 0.4 & 0.1 & 0.995 & 0.974 & Claude-Sonnet-4.5 \\
0.5 & 0.5 & 0.0 & 0.982 & 0.942 & Claude-Sonnet-4.5 \\
0.6 & 0.0 & 0.4 & 0.995 & 0.974 & Claude-Sonnet-4.5 \\
0.6 & 0.1 & 0.3 & 1.000 & 1.000 & Claude-Sonnet-4.5 \\
0.6 & 0.2 & 0.2 & 1.000 & 1.000 & Claude-Sonnet-4.5 \\
0.6 & 0.3 & 0.1 & 0.995 & 0.974 & Claude-Sonnet-4.5 \\
0.6 & 0.4 & 0.0 & 0.989 & 0.949 & Claude-Sonnet-4.5 \\
0.7 & 0.0 & 0.3 & 0.995 & 0.974 & Claude-Sonnet-4.5 \\
0.7 & 0.1 & 0.2 & 1.000 & 1.000 & Claude-Sonnet-4.5 \\
0.7 & 0.2 & 0.1 & 1.000 & 1.000 & Claude-Sonnet-4.5 \\
0.7 & 0.3 & 0.0 & 0.989 & 0.949 & Claude-Sonnet-4.5 \\
0.8 & 0.0 & 0.2 & 1.000 & 1.000 & Claude-Sonnet-4.5 \\
0.8 & 0.1 & 0.1 & 1.000 & 1.000 & Claude-Sonnet-4.5 \\
0.8 & 0.2 & 0.0 & 0.995 & 0.974 & Claude-Sonnet-4.5 \\
0.9 & 0.0 & 0.1 & 1.000 & 1.000 & Claude-Sonnet-4.5 \\
0.9 & 0.1 & 0.0 & 0.995 & 0.974 & Claude-Sonnet-4.5 \\
1.0 & 0.0 & 0.0 & 0.995 & 0.974 & Claude-Sonnet-4.5 \\
\end{longtable}
\normalsize

\begin{multicols}{2}
\raggedcolumns
\section{Qualitative Error Analysis}
\label{sec:qualitative_error_analysis}

Beyond aggregate scores, we conducted a qualitative inspection of failure cases to better understand why current data science agents underperform on complex tasks. We first analyze the judge rationales for 8,308 imperfect trajectories (i.e., those with final scores below 10) and categorize the cited capability gaps. Table~\ref{tab:error_overall} summarizes the most frequently cited capability gaps across all tasks, and Table~\ref{tab:error_domain} stratifies these results by domain.

We focus particularly on Computer Vision (CV) and Natural Language Processing (NLP), since these domains show consistent performance degradation in our main results.
Compared with structured tabular analysis, these tasks often require more complex data loading, custom preprocessing, specialized library usage, and model architecture design.
Based on manual inspection of representative failed trajectories, we identify four major failure modes.

\end{multicols}

\begin{table}[H]
\centering
\small
\begin{tabular}{lrr}
\toprule
\textbf{Capability-gap indicator} & \textbf{Count} & \textbf{Prevalence} \\
\midrule
Requirement coverage/completeness & 4,329 & 52.1\% \\
Visualization/result interpretation & 3,469 & 41.8\% \\
Data understanding/semantics & 3,256 & 39.2\% \\
Methodological/statistical reasoning & 2,931 & 35.3\% \\
Preprocessing/feature representation & 2,457 & 29.6\% \\
Training/validation/metric protocol & 1,616 & 19.5\% \\
Model architecture/configuration & 258 & 3.1\% \\
\bottomrule
\end{tabular}
\caption{Capability gaps cited in judge rationales for 8,308 imperfect trajectories. Categories overlap.}
\label{tab:error_overall}

\vspace{0.5em}
\scriptsize
\resizebox{\textwidth}{!}{
\begin{tabular}{lrrrrrrr}
\toprule
\textbf{Domain} & \textbf{$n$} & \textbf{Data} & \textbf{Method} & \textbf{Arch.} & \textbf{Train/eval} & \textbf{Complete} & \textbf{Visual} \\
\midrule
Data Analysis & 4,673 & 44.6 & 36.8 & 1.9 & 19.2 & 52.6 & 43.8 \\
Computer Vision & 356 & 37.9 & 36.0 & 16.9 & 38.2 & 49.4 & 33.1 \\
Natural Language Processing & 403 & 32.3 & 36.2 & 2.2 & 17.9 & 54.8 & 31.5 \\
Time Series & 748 & 37.7 & 29.3 & 1.2 & 16.8 & 51.9 & 44.4 \\
Clustering & 234 & 32.9 & 31.6 & 6.4 & 16.2 & 52.1 & 50.0 \\
\bottomrule
\end{tabular}}
\caption{Capability-gap prevalence (\%) by selected domain. Data: data understanding; Method: methodological/statistical reasoning; Arch.: architecture/configuration; Train/eval: training, validation, or metric protocol; Complete: requirement coverage; Visual: visualization/result interpretation.}
\label{tab:error_domain}
\end{table}

\begin{multicols}{2}
\raggedcolumns

\paragraph{Multi-step Code Execution Failures.}
The most frequent failure mode is the inability to complete critical pipeline stages over multiple turns.
In many cases, agents generate code that relies on hallucinated APIs, outdated library usage, or unsupported method calls.
Once runtime errors occur, the agent often fails to diagnose the root cause and instead makes local edits that do not resolve the underlying issue.
This leads to repeated execution failures and prevents the workflow from reaching the modeling or evaluation stage.
Such failures are especially common in tasks involving deep learning libraries, custom dataset classes, and non-standard data formats.

\paragraph{Incorrect Model Architecture Selection.}
Another common failure mode appears in model design, especially for Computer Vision tasks.
Agents sometimes speculate about complex neural network configurations without grounding these choices in the actual input data shape or task requirements.
For example, they may modify internal components of a pretrained CNN encoder, change classifier heads incorrectly, or assume incompatible feature dimensions.
These choices often lead to shape mismatch errors during the forward pass or training loop.
The problem is not merely a coding error; it reflects a lack of execution-aware architectural reasoning, where the agent fails to verify whether the proposed model is compatible with the data and training pipeline.

\paragraph{Data Loading and Preprocessing Issues.}
A significant number of failures occur at the earliest stage of the workflow.
Agents may fail to identify the correct files, parse nested directory structures, handle image or text datasets, or construct appropriate dataset loaders.
For NLP tasks, this may involve incorrect assumptions about text fields, labels, encodings, or tokenization formats.
For CV tasks, this may involve incorrect class-folder mapping, image transformations, or train-validation-test split handling.
When data loading or preprocessing fails, subsequent modeling steps cannot start, causing the entire task to fail even before substantive analysis begins.

\paragraph{Result Hallucination and Failure Masking.}
Beyond execution-level failures, we also observe cases where agents produce final reports that are not supported by executed code.
In these cases, the agent may confidently report specific accuracy values, metric scores, selected hyperparameters, or model configurations that were never successfully produced during execution.
This behavior masks the underlying failure and makes the final response appear more complete than it actually is.
Result hallucination is particularly concerning for data science agents because real-world data analysis often ends with a report-style deliverable.
If the report is not grounded in verifiable execution traces, users may be misled by plausible but unsupported conclusions.

\paragraph{Implications.}
These failure modes suggest that future data science agents need stronger mechanisms for execution-grounded reasoning.
First, agents should verify the validity of intermediate outputs before proceeding to downstream steps.
Second, model architecture choices should be checked against actual data shapes and task constraints.
Third, data loading and preprocessing should receive greater attention, especially for unstructured data.
Finally, final reports should be explicitly grounded in executed code and generated artifacts.
A promising direction is to introduce consistency checks between the notebook execution trace and the final report, ensuring that every reported metric, visualization, and conclusion can be traced back to actual computation.

\end{multicols}

\end{document}